\def\year{2018}\relax
\documentclass[letterpaper]{article} 
\usepackage{aaai18}  
\usepackage{times}  
\usepackage{helvet}  
\usepackage{courier}  
\usepackage{url}  
\usepackage{graphicx}  
\usepackage{subfigure}
\usepackage{multirow}
\usepackage{booktabs}
\usepackage{diagbox}
\usepackage{color}
\usepackage{bbm}
\usepackage{multicol}
\usepackage{blindtext}
\usepackage{float}
\usepackage{amsmath,amsfonts,amssymb}
\usepackage{algorithm}
\usepackage{algorithmic}
\usepackage{bm}
\DeclareMathOperator*{\softmax}{softmax}
\DeclareMathOperator*{\sigmoid}{sigmoid}

\graphicspath{{./}}

\begin{document}
%
\title{DiSAN: Directional Self-Attention Network for \\RNN/CNN-Free Language Understanding}
\author{
	Tao Shen\dag  \vspace{0mm} \\ {\bf \Large Jing Jiang\dag} 
	\And Tianyi Zhou\ddag  \vspace{0mm} \\ {\bf \Large Shirui Pan\dag}\\
	\And Guodong Long\dag  \vspace{0mm} \\ {\bf \Large Chengqi Zhang\dag}\vspace{3mm}
	\AND \vspace{-8mm}
	\\\dag Centre of Artificial Intelligence, FEIT, University of Technology Sydney
	\\\ddag Paul G. Allen School of Computer Science \& Engineering, University of Washington 
	\\ \texttt{tao.shen@student.uts.edu.au, tianyizh@uw.edu}  
	\\ \texttt{\{guodong.long, jing.jiang, shirui.pan, chengqi.zhang\}@uts.edu.au}
}

\maketitle
\begin{abstract}
Recurrent neural nets (RNN) and convolutional neural nets (CNN) are widely used on NLP tasks to capture the long-term and local dependencies, respectively. Attention mechanisms have recently attracted enormous interest due to their highly parallelizable computation, significantly less training time, and flexibility in modeling dependencies. We propose a novel attention mechanism in which the attention between elements from input sequence(s) is directional and multi-dimensional (i.e., feature-wise). A light-weight neural net, ``Directional Self-Attention Network (DiSAN)'', is then proposed to learn sentence embedding, based solely on the proposed attention without any RNN/CNN structure. DiSAN is only composed of a directional self-attention with temporal order encoded, followed by a multi-dimensional attention that compresses the sequence into a vector representation. Despite its simple form, DiSAN outperforms complicated RNN models on both prediction quality and time efficiency. It achieves the best test accuracy among all sentence encoding methods and improves the most recent best result by $1.02\%$ on the Stanford Natural Language Inference (SNLI) dataset, and shows state-of-the-art test accuracy on the Stanford Sentiment Treebank (SST), Multi-Genre natural language inference (MultiNLI), Sentences Involving Compositional Knowledge (SICK), Customer Review, MPQA, TREC question-type classification and Subjectivity (SUBJ)  datasets. 
\end{abstract}

\section{Introduction}

Context dependency plays a significant role in language understanding and provides critical information to natural language processing (NLP) tasks. For different tasks and data, researchers often switch between two types of deep neural network (DNN): recurrent neural network (RNN) with sequential architecture capturing long-range dependencies (e.g., long short-term memory (LSTM) \cite{hochreiter1997long} and gated recurrent unit (GRU) \cite{chung2014empirical}), and convolutional neural network (CNN) \cite{kim2014convolutional} whose hierarchical structure is good at extracting local or position-invariant features. However, which network to choose in practice is an open question, and the choice relies largely on the empirical knowledge.

Recent works have found that equipping RNN or CNN with an attention mechanism can achieve state-of-the-art performance on a large number of NLP tasks, including neural machine translation \cite{bahdanau2015neural,luong2015effective}, natural language inference \cite{liu2016learning}, conversation generation \cite{shang2015neural}, question answering \cite{hermann2015teaching,sukhbaatar2015end}, machine reading comprehension \cite{seo2017bidirectional}, and sentiment analysis \cite{kokkinos2017structural}. The attention uses a hidden layer to compute a categorical distribution over elements from the input sequence  to reflect their importance weights. It allows RNN/CNN to maintain a variable-length memory, so that elements from the input sequence can be selected by their importance/relevance and merged into the output. In contrast to RNN and CNN, the attention mechanism is trained to capture the dependencies that make significant contributions to the task, regardless of the distance between the elements in the sequence. It can thus provide complementary information to the distance-aware dependencies modeled by RNN/CNN. In addition, computing attention only requires matrix multiplication, which is highly parallelizable compared to the sequential computation of RNN.

In a very recent work \cite{vaswani2017attention}, an attention mechanism is solely used to construct a sequence to sequence (seq2seq) model that achieves a state-of-the-art quality score on the neural machine translation (NMT) task. The seq2seq model, ``Transformer'', has an encoder-decoder structure that is only composed of stacked attention networks, without using either recurrence or convolution. The proposed attention, ``multi-head attention'', projects the input sequence to multiple subspaces, then applies scaled dot-product attention to its representation in each subspace, and lastly concatenates their output. By doing this, it can combine different attentions from multiple subspaces. This mechanism is used in Transformer to compute both the context-aware features inside the encoder/decoder and the bottleneck features between them.

The attention mechanism has more flexibility in sequence length than RNN/CNN, and is more task/data-driven when modeling dependencies. Unlike sequential models, its computation can be easily and significantly accelerated by existing distributed/parallel computing schemes. However, to the best of our knowledge, a neural net entirely based on attention has not been designed for other NLP tasks except NMT, especially those that cannot be cast into a seq2seq problem. Compared to RNN, a disadvantage of most attention mechanisms is that the temporal order information is lost, which however might be important to the task. This explains why positional encoding is applied to the sequence before being processed by the attention in Transformer. How to model order information within an attention is still an open problem.

The goal of this paper is to develop a unified and RNN/CNN-free attention network that can be generally utilized to learn the sentence encoding model for different NLP tasks, such as natural language inference, sentiment analysis, sentence classification and semantic relatedness. We focus on the sentence encoding model because it is a basic module of most DNNs used in the NLP literature. 

We propose a novel attention mechanism that differs from previous ones in that it is 1) multi-dimensional: the attention w.r.t. each pair of elements from the source(s) is a vector, where each entry is the attention computed on each feature; and 2) directional: it uses one or multiple positional masks to model the asymmetric attention between two elements. We compute feature-wise attention since each element in a sequence is usually represented by a vector, e.g., word/character embedding \cite{kim2016character}, and attention on different features can contain different information about dependency, thus to handle the variation of contexts around the same word. We apply positional masks to attention distribution since they can easily encode prior structure knowledge such as temporal order and dependency parsing. This design mitigates the weakness of attention in modeling order information, and takes full advantage of parallel computing. 

We then build a light-weight and RNN/CNN-free neural network, ``Directional Self-Attention Network (DiSAN)'', for sentence encoding. This network relies entirely on the proposed attentions and does not use any RNN/CNN structure. In DiSAN, the input sequence is processed by directional (forward and backward) self-attentions to model context dependency and produce context-aware representations for all tokens. Then, a multi-dimensional attention computes a vector representation of the entire sequence, which can be passed into a classification/regression module to compute the final prediction for a particular task. Unlike Transformer, neither stacking of attention blocks nor an encoder-decoder structure is required. The simple architecture of DiSAN leads to fewer parameters, less computation and easier parallelization. 

In experiments\footnote{Codes and pre-trained models for experiments can be found at \texttt{https://github.com/taoshen58/DiSAN}}, we compare DiSAN with the currently popular methods on various NLP tasks, e.g., natural language inference, sentiment analysis, sentence classification, etc. DiSAN achieves the highest test accuracy on the Stanford Natural Language Inference (SNLI) dataset among sentence-encoding models and improves the currently best result by $1.02\%$. It also shows the state-of-the-art performance on the Stanford Sentiment Treebank (SST),  Multi-Genre natural language inference (MultiNLI), SICK, Customer Review, MPQA, SUBJ and TREC question-type classification datasets. Meanwhile, it has fewer parameters and exhibits much higher computation efficiency than the models it outperforms, e.g., LSTM and tree-based models. 

\textbf{Annotation:} 1) Lowercase denotes a vector; 2)  bold lowercase denotes a sequence of vectors (stored as a matrix); and 3) uppercase denotes a matrix or a tensor.

\section{Background}

\subsection{Sentence Encoding}

In the pipeline of NLP tasks, a sentence is denoted by a sequence of discrete tokens (e.g., words or characters) $\bm{v} = [v_1, v_2, \dots, v_n]$, where $v_i$ could be a one-hot vector whose dimension length equals the number of distinct tokens $N$. A pre-trained token embedding (e.g., word2vec \cite{mikolov2013distributed} or GloVe \cite{pennington2014glove}) is applied to $\bm{v}$ and transforms all discrete tokens to a sequence of low-dimensional dense vector representations $\bm{x}=[x_1, x_2, \dots, x_n]$ with $x_i\in\mathbb R^{d_e }$. This pre-process can be written as $\bm{x} = W^{(e)} \bm{v}$, where word embedding weight matrix $W^{(e)}\in\mathbb R^{ d_e \times N}$ and $\bm{x}\in\mathbb R^{d_e \times n}$.

Most DNN sentence-encoding models for NLP tasks take $\bm{x}$ as the input and further generate a vector representation $u_i$ for each $x_i$ by context fusion. Then a sentence encoding is obtained by mapping the sequence $\bm{u}=[u_1,u_2,\dots, u_n]$ to a single vector $s \in \mathbb R^d$, which is used as a compact encoding of the entire sentence in NLP problems.

\subsection{Attention}

The attention is proposed to compute an alignment score between elements from two sources. In particular, given the token embeddings of a source sequence $\bm{x}=[x_1, x_2, \dots, x_n]$ and the vector representation of a query $q$, attention computes the alignment score between $x_i$ and $q$ by a compatibility function $f(x_i, q)$, which measures the dependency between $x_i$ and $q$, or the attention of $q$ to $x_i$. A $\softmax$ function then transforms the scores $[f(x_i, q)]_{i=1}^{n}$ to a probability distribution $p(z|\bm{x}, q)$ by normalizing over all the $n$ tokens of $\bm{x}$. Here $z$ is an indicator of which token in $\bm{x}$ is important to $q$ on a specific task. That is, large $p(z=i|\bm{x}, q)$ means $x_i$ contributes important information to $q$. The above process can be summarized by the following equations.
\begin{align}
&a = \left[f(x_i, q)\right]_{i=1}^n \label{eq_tra_attn_1},\\
&p(z|\bm{x}, q) = \softmax(a) \label{eq_tra_attn_2}.
\end{align}
Specifically,
\begin{equation}\label{equ:attention_prob}
p(z=i|\bm{x}, q) = \frac{\exp(f(x_i, q))}{\sum_{i=1}^n\exp(f(x_i, q))}.
\end{equation}
The output of this attention mechanism is a weighted sum of the embeddings for all tokens in $\bm{x}$, where the weights are given by $p(z|\bm{x},q)$. It places large weights on the tokens important to $q$, and can be written as the expectation of a token sampled according to its importance, i.e.,
\begin{equation}\label{equ:attention_output}
s = \sum_{i=1}^n p(z=i|\bm{x},q)x_i=\mathbb E_{i\sim p(z|\bm{x},q)}(x_i),
\end{equation}
where $s \in \mathbb{R}^{d_e}$ can be used as the sentence encoding of $\bm{x}$.

\begin{figure}[htbp]
	\centering
	\subfigure[]{
		\label{fig:tra_attn:1a}
		\includegraphics[width=0.22\textwidth]{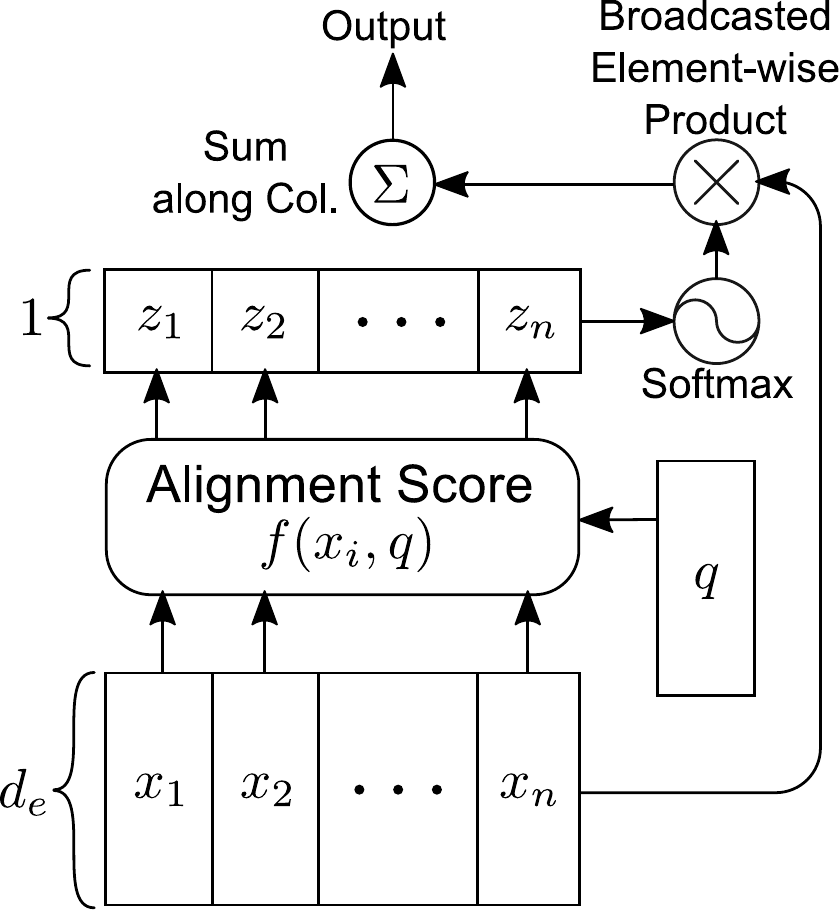}}
	\subfigure[]{
		\label{fig:mul_attn:1b} 
		\includegraphics[width=0.22\textwidth]{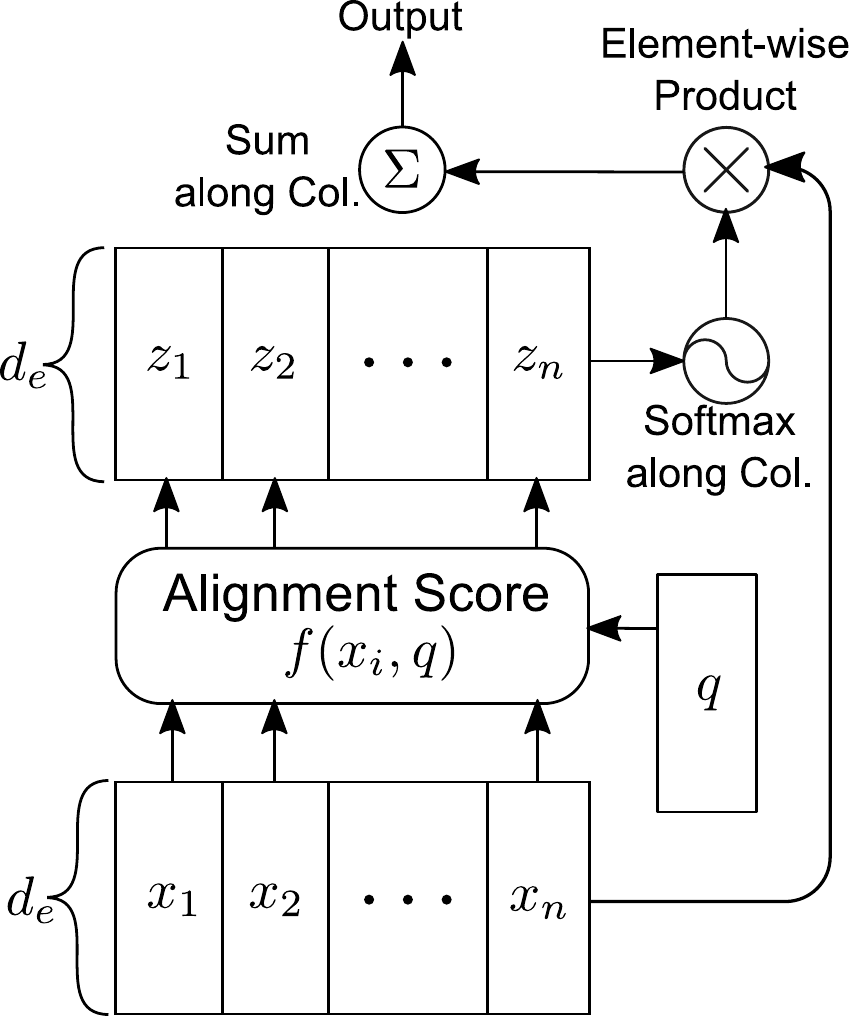}}
	\caption{(a) Traditional (additive/multiplicative) attention and (b) multi-dimensional attention. $z_i$ denotes alignment score $f(x_i, q)$, which is a scalar in (a) but a vector in (b).}
	\label{fig:attns} 
	\centering
\end{figure}
Additive attention (or multi-layer perceptron attention) \cite{bahdanau2015neural,shang2015neural} and multiplicative attention (or dot-product attention) \cite{vaswani2017attention,sukhbaatar2015end,rush2015neural} are the two most commonly used attention mechanisms. They share the same and unified form of attention introduced above, but are different in the compatibility function $f(x_i, q)$. Additive attention is associated with
\begin{equation}\label{equ:add_attention}
f(x_i, q)=w^T\sigma(W^{(1)}x_i+W^{(2)}q),
\end{equation}
where $\sigma(\cdot)$ is an activation function and $w \in \mathbb{R}^{d_e}$ is a weight vector. Multiplicative attention uses inner product or cosine similarity for $f(x_i, q)$, i.e.,
\begin{equation}
f(x_i, q)=\left\langle W^{(1)}x_i, W^{(2)}q\right\rangle.
\end{equation}
In practice, additive attention often outperforms multiplicative one in prediction quality, but the latter is faster and more memory-efficient due to optimized matrix multiplication. 

\subsection{Self-Attention}

Self-Attention is a special case of the attention mechanism introduced above. It replaces $q$ with a token embedding $x_j$ from the source input itself. It relates elements at different positions from a single sequence by computing the attention between each pair of tokens, $x_i$ and $x_j$. It is very expressive and flexible for both long-range and local dependencies, which used to be respectively modeled by RNN and CNN. Moreover, it has much faster computation speed and fewer parameters than RNN. In recent works, we have already witnessed its success across a variety of NLP tasks, such as reading comprehension \cite{hu2017mnemonic} and neural machine translation \cite{vaswani2017attention}.

\section{Two Proposed Attention Mechanisms}\label{sec:2attentions}

In this section, we introduce two novel attention mechanisms, multi-dimensional attention in Section \ref{sec:multi_dim_attn} (with two extensions to self-attention in Section \ref{sec:2extensions}) and directional self-attention in Section \ref{sec:self_attn}. They are the main components of DiSAN and may be of independent interest to other neural nets for other NLP problems in which an attention is needed.

\subsection{Multi-dimensional Attention}\label{sec:multi_dim_attn}

Multi-dimensional attention is a natural extension of additive attention (or MLP attention) at the feature level. Instead of computing a single scalar score $f(x_i, q)$ for each token $x_i$ as shown in Eq.(\ref{equ:add_attention}), multi-dimensional attention computes a feature-wise score vector for $x_i$ by replacing weight vector $w$ in Eq.(\ref{equ:add_attention}) with a matrix $W$, i.e.,
\begin{equation}\label{equ:md_attention}
f(x_i, q)=W^T\sigma\left(W^{(1)}x_i+W^{(2)}q\right),
\end{equation}
where $f(x_i, q) \in \mathbb{R}^{d_e}$ is a vector with the same length as $x_i$, and all the weight matrices $W,W^{(1)},W^{(2)}\in\mathbb R^{d_e\times d_e}$. We further add two bias terms to the parts in and out activation $\sigma (\cdot)$, i.e.,
\begin{equation}\label{equ:md_attention_bias}
f(x_i, q)=W^T\sigma\left(W^{(1)}x_i+W^{(2)}q+b^{(1)}\right)+b.
\end{equation}
We then compute a categorical distribution $p(z_k|\bm{x}, q)$ over all the $n$ tokens for each feature $k\in[d_e]$. A large $p(z_k=i|\bm{x}, q)$ means that feature $k$ of token $i$ is important to $q$. 

We apply the same procedure Eq.(\ref{eq_tra_attn_1})-(\ref{equ:attention_prob}) in traditional attention to the $k^{th}$ dimension of $f(x_i, q)$. In particular, for each feature $k\in[d_e]$, we replace $f(x_i, q)$ with $[f(x_i, q)]_k$, and change $z$ to $z_k$ in Eq.(\ref{eq_tra_attn_1})-(\ref{equ:attention_prob}). 
Now each feature $k$ in each token $i$ has an importance weight $P_{ki}\triangleq p(z_k=i|\bm{x}, q)$. The output $s$ can be written as
\begin{equation}
s = \left[\sum\nolimits_{i=1}^n P_{ki}\bm{x}_{ki}\right]_{k=1}^{d_e}=\left[\mathbb E_{i\sim p(z_k|\bm{x},q)} (\bm{x}_{ki})\right]_{k=1}^{d_e}.
\end{equation}

We give an illustration of traditional attention and multi-dimensional attention in Figure \ref{fig:attns}. In the rest of this paper, we will ignore the subscript $k$ which indexes feature dimension for simplification if no confusion is possible. Hence, the output $s$ can be written as an element-wise product $s = \sum_{i=1}^n P_{\cdot i} \odot x_i$

\textbf{Remark}: The word embedding usually suffers from the polysemy in natural language. Since traditional attention computes a single importance score for each word based on the word embedding, it cannot distinguish the meanings of the same word in different contexts. Multi-dimensional attention, however, computes a score for each feature of each word, so it can select the features that can best describe the word's specific meaning in any given context, and include this information in the sentence encoding output $s$.

\subsection{Two types of Multi-dimensional Self-attention}\label{sec:2extensions}

When extending multi-dimension to self-attentions, we have two variants of multi-dimensional attention. The first one, called multi-dimensional ``\textbf{token2token}'' self-attention, explores the dependency between $x_i$ and $x_j$ from the same source $\bm{x}$, and generates context-aware coding for each element. It replaces $q$ with $x_j$ in Eq.(\ref{equ:md_attention_bias}), i.e.,
\begin{equation}\label{equ:md_attention_self_token2token}
f(x_i, x_j)=W^T\sigma\left(W^{(1)}x_i+W^{(2)}x_j+b^{(1)}\right)+b.
\end{equation}
Similar to $P$ in vanilla multi-dimensional attention, we compute a probability matrix $P^{j} \in \mathbb R ^{d_e\times n}$ for each $x_j$ such that $P^{j}_{ki}\triangleq p(z_k=i|\bm{x}, x_j)$.
The output for $x_j$ is
\begin{equation}\label{equ:sentence_encoding_self_token2token}
s_j=\sum_{i=1}^nP_{\cdot i}^{j} \odot x_i
\end{equation}
The output of token2token self-attention for all elements from $\bm{x}$ is $\bm{s} = [s_1, s_2, \dots, s_n] \in \mathbb R^{d_e \times n}$.

The second one, multi-dimensional ``\textbf{source2token}'' self-attention, explores the dependency between $x_i$ and the entire sequence $\bm{x}$, and compresses the sequence $\bm{x}$ into a vector. It removes $q$ from Eq.(\ref{equ:md_attention_bias}), i.e.,
\begin{equation}\label{equ:md_attention_self_source2token}
f(x_i)=W^T\sigma\left(W^{(1)}x_i+b^{(1)}\right)+b.
\end{equation}
The probability matrix is defined as $P_{ki}\triangleq p(z_k=i|\bm{x})$ and is computed in the same way as $P$ in vanilla multi-dimensional attention. The output $s$ is also same, i.e.,
\begin{equation}\label{equ:sentence_encoding_self_source2token}
s=\sum_{i=1}^{n} P_{\cdot i} \odot x_i
\end{equation}

We will use these two types (i.e., token2token and source2token) of multi-dimensional self-attention in different parts of our sentence encoding model, DiSAN.

\subsection{Directional Self-Attention} \label{sec:self_attn}

\begin{figure}[t]
	\includegraphics[width=0.45\textwidth]{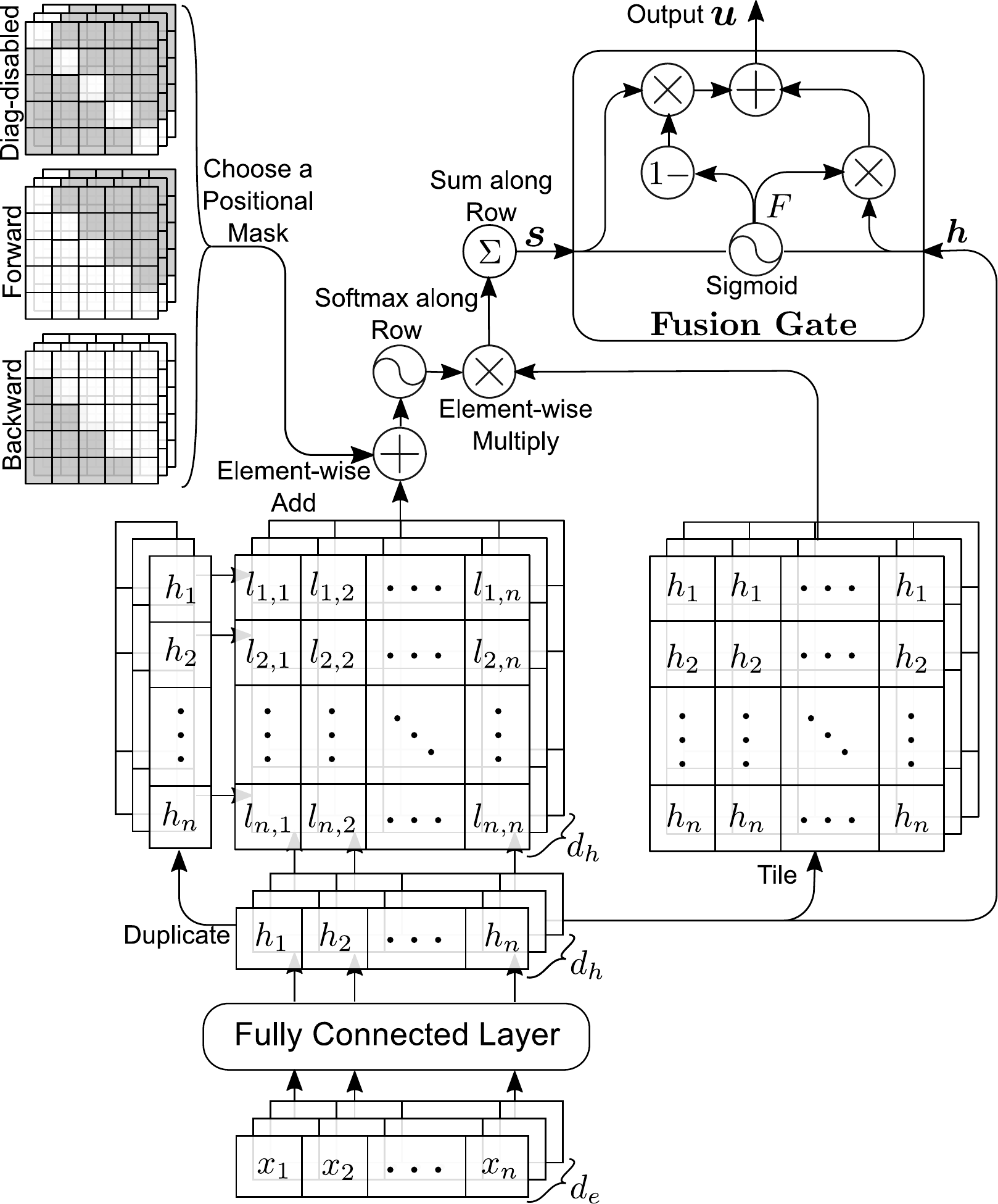}
	\caption{Directional self-attention (DiSA) mechanism. Here, we use $l_{i, j}$ to denote $f(h_i, h_j)$ in Eq. (\ref{equ:ds_attention_self_token2token_mask}).}
	\label{fig:self_attn} 
	\centering
\end{figure}

Directional self-attention (DiSA) is composed of a fully connected layer whose input is the token embeddings $\bm{x}$, a ``masked'' multi-dimensional token2token self-attention block to explore the dependency and temporal order, and a fusion gate to combine the output and input of the attention block. Its structure is shown in Figure \ref{fig:self_attn}. It can be used as either a neural net or a module to compose a large network.

In DiSA, we first transform the input sequence $\bm{x}=[x_1,x_2,\dots, x_n]$ to a sequence of hidden state $\bm{h}=[h_1, h_2, \dots, h_n]$ by a fully connected layer, i.e.,
\begin{equation}\label{equ:fc_output}
\bm{h}=\sigma_h\left(W^{(h)}\bm{x}+b^{(h)}\right),
\end{equation}
where $\bm{x}\in\mathbb R^{d_e\times n}$, $\bm{h}\in\mathbb R^{d_h\times n}$, $W^{(h)}$ and $b^{(h)}$ are the learnable parameters, and $\sigma_h(\cdot)$ is an activation function.

We then apply multi-dimensional token2token self-attention to $\bm{h}$, and generate context-aware vector representations $\bm{s}$ for all elements from the input sequence. We make two modifications to Eq.(\ref{equ:md_attention_self_token2token}) to reduce the number of parameters and make the attention directional.

First, we set $W$ in Eq.(\ref{equ:md_attention_self_token2token}) to a scalar $c$ and divide the part in $\sigma (\cdot)$ by $c$, and we use $\tanh(\cdot)$ for $\sigma(\cdot)$, which reduces the number of parameters. In experiments, we always set $c=5$, and obtain stable output. 

Second, we apply a positional mask to Eq.(\ref{equ:md_attention_self_token2token}), so the attention between two elements can be asymmetric. Given a mask $M\in\{0, -\infty\}^{n\times n}$, we set bias $b$ to a constant vector $M_{ij} \bm{1}$ in Eq.(\ref{equ:md_attention_self_token2token}), where $\bm{1}$ is an all-one vector. Hence, Eq.(\ref{equ:md_attention_self_token2token}) is modified to
\begin{align}\label{equ:ds_attention_self_token2token_mask}
\notag &f(h_i, h_j)=\\
&c\cdot\tanh\left([W^{(1)}h_i+W^{(2)}h_j+b^{(1)}]/c\right)
+M_{ij} \bm{1}.
\end{align}

To see why a mask can encode directional information, let us consider a case in which $M_{ij}=-\infty$ and $M_{ji}=0$, which results in $[f(h_i, h_j)]_k=-\infty$ and unchanged $[f(h_j, h_i)]_k$. Since the probability $p(z_k=i|\bm{x},x_j)$ is computed by $\softmax$, $[f(h_i, h_j)]_k=-\infty$ leads to $p(z_k=i|\bm{x},x_j)=0$. This means that there is no attention of $x_j$ to $x_i$ on feature $k$. On the contrary, we have $p(z_k=j|\bm{x},x_i)>0$, which means that attention of $x_i$ to $x_j$ exists on feature $k$. Therefore, prior structure knowledge such as temporal order and dependency parsing can be easily encoded by the mask, and explored in generating sentence encoding. This is an important feature of DiSA that previous attention mechanisms do not have.

\begin{figure}[htbp]
	\centering
	\subfigure[Diag-disabled mask]{
		\label{fig:mask_diag:2a}
		\includegraphics[width=0.2\textwidth]{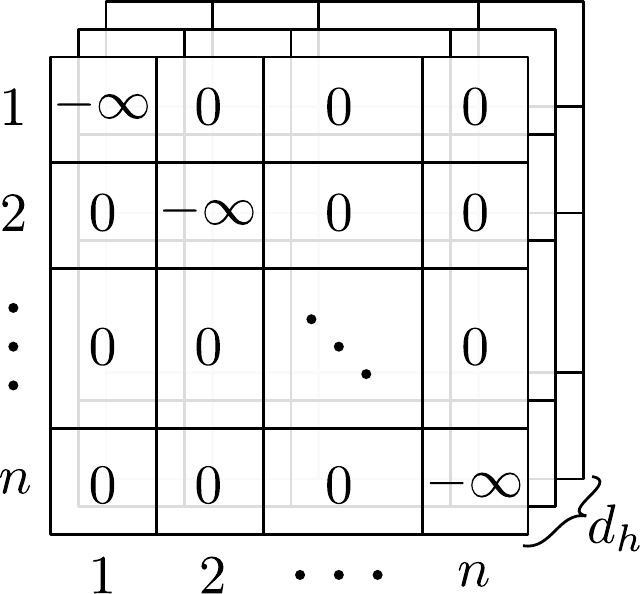}}\\
	\subfigure[Forward mask]{
		\label{fig:mask_fw:2b} 
		\includegraphics[width=0.2\textwidth]{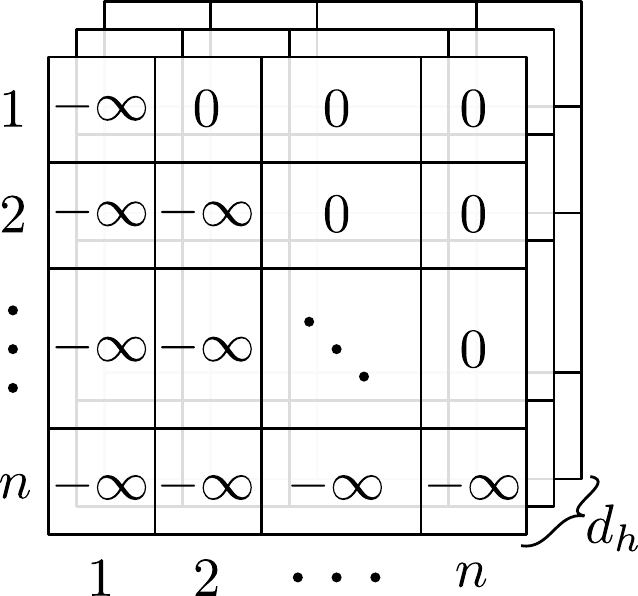}}
	\subfigure[Backward mask]{
		\label{fig:mask_bw:2c} 
		\includegraphics[width=0.2\textwidth]{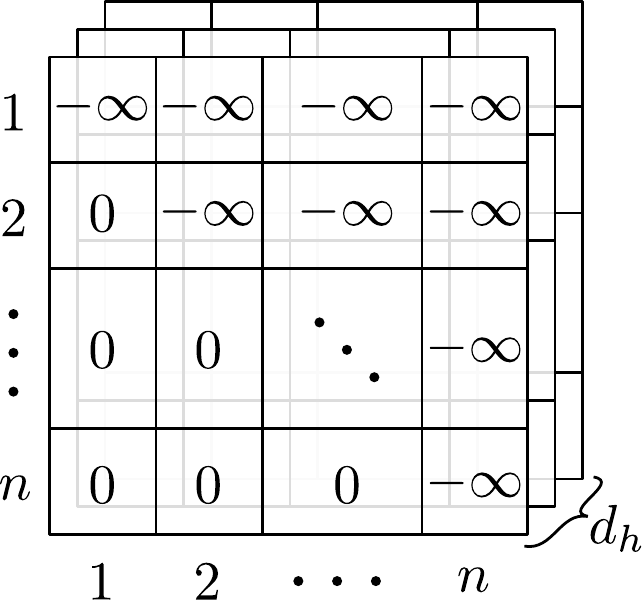}}
	\caption{Three positional masks: (a) is the diag-disabled mask $M^{diag}$;  (b) and (c) are forward mask $M^{fw}$ and backward mask $M^{bw}$, respectively.}
	\label{fig:masks} 
	\centering
\end{figure}

For self-attention, we usually need to disable the attention of each token to itself \cite{hu2017mnemonic}. This is the same as applying a diagonal-disabled (i.e., diag-disabled) mask such that
\begin{equation} \label{equ:diag_disable_mask}
M_{ij}^{diag} = \left\{\begin{matrix}
0,& i \neq j\\
- \infty,& i = j
\end{matrix}\right.
\end{equation}

Moreover, we can use masks to encode temporal order information into attention output. In this paper, we use two masks, i.e., forward mask $M^{fw}$ and backward mask $M^{bw}$,
\begin{align}
&M_{ij}^{fw} = \left\{\begin{matrix}
0,& i < j \\ 
- \infty,& otherwise
\end{matrix}\right. \label{equ:fw_mask}\\
&M_{ij}^{bw} = \left\{\begin{matrix}
0,& i > j \\ 
- \infty,& otherwise
\end{matrix}\right. \label{equ:bw_mask}
\end{align}
In forward mask $M^{fw}$, there is the only attention of later token $j$ to early token $i$, and vice versa in backward mask. We show these three positional masks in Figure \ref{fig:masks}.

Given input sequence $\bm{x}$ and a mask $M$, we compute $f(x_i, x_j)$ according to Eq.(\ref{equ:ds_attention_self_token2token_mask}), and follow the standard procedure of multi-dimensional token2token self-attention to compute the probability matrix $P^{j}$ for each $j\in[n]$. Each output $s_j$ in $\bm{s}$ is computed as in Eq.(\ref{equ:sentence_encoding_self_token2token}).

The final output $\bm{u}\in \mathbb R^{d_h\times n}$ of DiSA is obtained by combining the output $\bm{s}$ and the input $\bm{h}$ of the masked multi-dimensional token2token self-attention block. This yields a temporal order encoded and context-aware vector representation for each element/token. The combination is accomplished by a dimension-wise fusion gate, i.e.,
\begin{align}
&F = \sigmoid\left(W^{(f1)}\bm{s} + W^{(f2)}\bm{h} + b^{(f)}\right) \label{eq_self_attn_6} \\
&\bm{u} =F \odot \bm{h} +  (1 - F)  \odot \bm{s} \label{eq_self_attn_7}
\end{align}
where $W^{(f1)},W^{(f2)} \in \mathbb R^{d_h \times d_h}$ and $b^{(f)} \in \mathbb R^{d_h}$ are the learnable parameters of the fusion gate. 

\section{Directional Self-Attention Network}

\begin{figure}[htbp]
	\centering
	\includegraphics[width=0.35\textwidth]{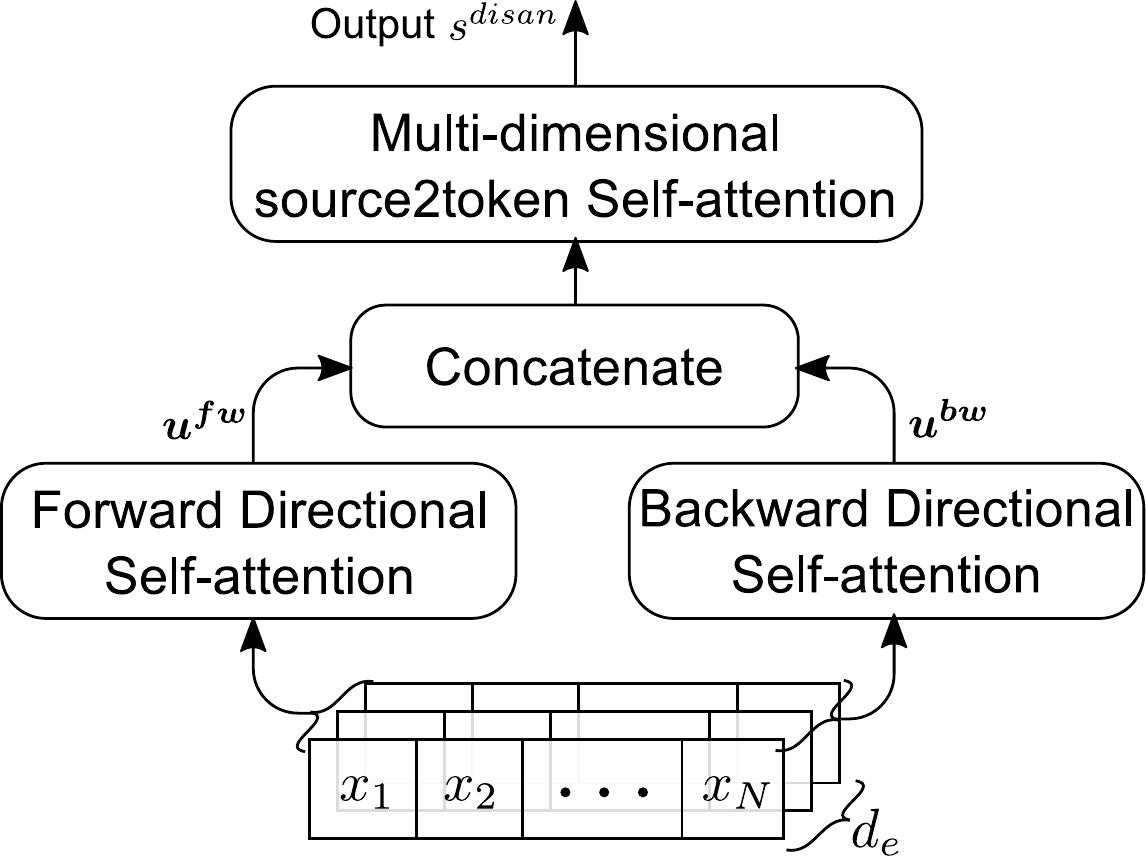}
	\caption{Directional self-attention network (DiSAN)}
	\label{fig:model} 
\end{figure}

We propose a light-weight network, ``Directional Self-Attention Network (DiSAN)'', for sentence encoding. Its architecture is shown in Figure \ref{fig:model}. 

Given an input sequence of token embedding $\bm{x}$, DiSAN firstly applies two parameter-untied DiSA blocks with forward mask $M^{fw}$ Eq.(\ref{equ:fw_mask}) and $M^{bw}$ Eq.(\ref{equ:bw_mask}), respectively. The feed-forward procedure is given in Eq.(\ref{equ:fc_output})-(\ref{equ:ds_attention_self_token2token_mask}) and Eq.(\ref{eq_self_attn_6})-(\ref{eq_self_attn_7}). Their outputs are denoted by $\bm{u^{fw}}, \bm{u^{bw}}\in\mathbb R^{d_h\times n}$. We concatenate them vertically as $[\bm{u^{fw}}; \bm{u^{bw}}] \in \mathbb{R}^{2d_h \times n} $, and use this concatenated output as input to a multi-dimensional source2token self-attention block, whose output $s^{disan} \in \mathbb{R}^{2d_h}$ computed by Eq.(\ref{equ:md_attention_self_source2token})-(\ref{equ:sentence_encoding_self_source2token}) is the final sentence encoding result of DiSAN.

\textbf{Remark}: In DiSAN, forward/backward DiSA blocks work as context fusion layers. And the multi-dimensional source2token self-attention compresses the sequence into a single vector. The idea of using both forward and backward attentions is inspired by Bi-directional LSTM (Bi-LSTM) \cite{graves2013hybrid}, in which forward and backward LSTMs are used to encode long-range dependency from different directions. In Bi-LSTM, LSTM combines the context-aware output with the input by multi-gate. The fusion gate used in DiSA shares the similar motivation. However, DiSAN has fewer parameters, simpler structure and better efficiency.

\section{Experiments}

\begin{table*}[htbp] \small
	\centering
	\begin{tabular}{@{}lcccc@{}}
		\toprule
		\textbf{Model Name}& \textbf{$\boldsymbol{|\theta|}$} & \textbf{T(s)/epoch} & \textbf{Train Accu(\%)} & \textbf{Test Accu(\%)} \\ \midrule
		Unlexicalized features \cite{bowman2015snli}&            &               &                49.4&               50.4\\
		+ Unigram and bigram features \cite{bowman2015snli}&            &               &                99.7&               78.2\\ \midrule
		100D LSTM encoders \cite{bowman2015snli}&            0.2m&               &                84.8&               77.6\\ 
		300D LSTM encoders \cite{bowman2016fast}&            3.0m&               &                83.9&               80.6\\ 
		1024D GRU encoders \cite{vendrov2016order}&            15m&               &                98.8&               81.4\\ 
		300D Tree-based CNN encoders \cite{mou2016natural}&            3.5m&               &                83.3&               82.1\\ 
		300D SPINN-PI encoders \cite{bowman2016fast}&            3.7m&               &                89.2&               83.2\\ 
		600D Bi-LSTM encoders \cite{liu2016learning}&            2.0m&               &                86.4&               83.3\\ 
		300D NTI-SLSTM-LSTM encoders \cite{munkhdalai2017neural}&            4.0m&               &                82.5&               83.4\\ 
		600D Bi-LSTM encoders+intra-attention \cite{liu2016learning}&            2.8m&               &                84.5&               84.2\\ 
		300D NSE encoders \cite{munkhdalai2016neural_2}&            3.0m&               &                86.2&               84.6\\ \midrule
		Word Embedding with additive attention&            0.45m&               216&                82.39&               79.81\\ 
		Word Embedding with s2t self-attention&            0.54m&               261&                86.22&               83.12\\ 
		Multi-head with s2t self-attention&            1.98m&               345&                89.58&               84.17\\ 
		Bi-LSTM with s2t self-attention&            2.88m&               2080&                90.39&               84.98\\ 
		DiSAN without directions&            2.35m&               592&                90.18&               84.66\\ \midrule
		Directional self-attention network  (DiSAN)&            2.35m&               587&                91.08&               \textbf{85.62}\\ \bottomrule
	\end{tabular}
	\caption{Experimental results for different methods on SNLI. \textbf{$\boldsymbol{|\theta|}$}: the number of parameters (excluding word embedding part). \textbf{T(s)/epoch}: average time (second) per epoch. \textbf{ Train Accu(\%)} and \textbf{Test Accu(\%)}: the accuracy on training and test set.}
	\label{table:snli_exps}
\end{table*}

In this section, we first apply DiSAN to natural language inference and sentiment analysis tasks.  DiSAN achieves the state-of-the-art performance and significantly better efficiency than other baseline methods on benchmark datasets for both tasks. We also conduct experiments on other NLP tasks and DiSAN also achieves state-of-the-art performance.

\textbf{Training Setup}: We use cross-entropy loss plus L2 regularization penalty as optimization objective. We minimize it by Adadelta \cite{zeiler2012adadelta} (an optimizer of mini-batch SGD) with batch size of $64$. We use Adadelta rather than Adam \cite{kingma2014adam} because in our experiments, DiSAN optimized by Adadelta can achieve more stable performance than Adam optimized one. Initial learning rate is set to $0.5$. All weight matrices are initialized by Glorot Initialization \cite{glorot2010understanding}, and the biases are initialized with $0$.
We initialize the word embedding in $\bm{x}$ by 300D GloVe 6B pre-trained vectors \cite{pennington2014glove}. The Out-of-Vocabulary words in training set are randomly initialized by uniform distribution between $(-0.05, 0.05)$. The word embeddings are fine-tuned during the training phrase.
We use Dropout \cite{srivastava2014dropout} with keep probability $0.75$ for language inference and $0.8$ for sentiment analysis. The L2 regularization decay factors $\gamma$ are $5\times10^{-5}$ and $10^{-4}$ for language inference and sentiment analysis, respectively. Note that the dropout keep probability and $\gamma$ varies with the scale of corresponding dataset. Hidden units number $d_h$ is set to $300$. Activation functions $\sigma(\cdot)$ are \textit{ELU} (exponential linear unit) \cite{clevert2016fast} if not specified. All models are implemented with TensorFlow\footnote{https://www.tensorflow.org} and run on single Nvidia GTX 1080Ti graphic card.

\subsection{Natural Language Inference} \label{sec:exp_nli}

The goal of Natural Language Inference (NLI) is to reason the semantic relationship between a premise sentence and a corresponding hypothesis sentence. The possible relationship could be \textit{entailment}, \textit{neutral} or \textit{contradiction}. We compare different models on a widely used benchmark, Stanford Natural Language Inference (SNLI)\footnote{https://nlp.stanford.edu/projects/snli/} \cite{bowman2015snli} dataset, which consists of 549,367/9,842/9,824 (train/dev/test) premise-hypothesis pairs with labels.

Following the standard procedure in \citeauthor{bowman2016fast} \shortcite{bowman2016fast}, we launch two sentence encoding models (e.g., DiSAN) with tied parameters for the premise sentence and hypothesis sentence, respectively. Given the output encoding $s^{p}$ for the premise and $s^{h}$ for the hypothesis, the representation of relationship is the concatenation of $s^{p}$, $s^{h}$, $s^{p} - s^{h}$ and $s^{p} \odot s^{h}$, which is fed into a 300D fully connected layer and then a $3$-unit output layer with $\softmax$ to compute a probability distribution over the three types of relationship.

For thorough comparison, besides the neural nets proposed in previous works of NLI, we implement five extra neural net baselines to compare with DiSAN. They help us to analyze the improvement contributed by each part of DiSAN and to verify that the two attention mechanisms proposed in Section \ref{sec:2attentions} can improve other networks.
\begin{itemize}
	\item \textbf{Word Embedding with additive attention}.
	\item \textbf{Word Embedding with s2t self-attention}: DiSAN with DiSA blocks removed.
	\item \textbf{Multi-head with s2t self-attention}: Multi-head attention \cite{vaswani2017attention} ($8$ heads, each has $75$ hidden units) with source2token self-attention. The positional encoding method used in \citeauthor{vaswani2017attention} \shortcite{vaswani2017attention} is applied to the input sequence to encode temporal information. We find our experiments show that multi-head attention is sensitive to hyperparameters, so we adjust keep probability of dropout from $0.7$ to $0.9$ with step $0.05$ and report the best result.
	\item \textbf{Bi-LSTM with s2t self-attention}: a multi-dimensional source2token self-attention block is applied to the output of Bi-LSTM (300D forward  + 300D backward LSTMs).
	\item \textbf{DiSAN without directions}: DiSAN with the forward/backward masks $M^{fw}$ and $M^{bw}$ replaced with two diag-disabled masks $M^{diag}$, i.e., DiSAN without forward/backward order information.
\end{itemize}

Compared to the results from the official leaderboard of SNLI in Table \ref{table:snli_exps}, DiSAN outperforms previous works and improves the best latest test accuracy (achieved by a memory-based NSE encoder network) by a remarkable margin of $1.02\%$. DiSAN surpasses the RNN/CNN based models with more complicated architecture and more parameters by large margins, e.g., $+2.32\%$ to Bi-LSTM, $+1.42\%$ to Bi-LSTM with additive attention. It even outperforms models with the assistance of a semantic parsing tree, e.g., $+3.52\%$ to Tree-based CNN, $+2.42\%$ to SPINN-PI.

In the results of the five baseline methods and DiSAN at the bottom of Table \ref{table:snli_exps}, we demonstrate that making attention multi-dimensional (feature-wise) or directional brings substantial improvement to different neural nets. First, a comparison between the first two models shows that changing token-wise attention to multi-dimensional/feature-wise attention leads to $3.31\%$ improvement on a word embedding based model. Also, a comparison between the third baseline and DiSAN shows that DiSAN can substantially outperform multi-head attention by $1.45\%$. Moreover, a comparison between the forth baseline and DiSAN shows that the DiSA block can even outperform Bi-LSTM layer in context encoding, improving test accuracy by $0.64\%$. A comparison between the fifth baseline and DiSAN shows that directional self-attention with forward and backward masks (with temporal order encoded) can bring $0.96\% $ improvement.

Additional advantages of DiSAN shown in Table \ref{table:snli_exps} are its fewer parameters and compelling time efficiency. It is $\times3$ faster than widely used Bi-LSTM model. Compared to other models with competitive performance, e.g., 600D Bi-LSTM encoders with intra-attention (2.8M), 300D NSE encoders (3.0M) and 600D Bi-LSTM encoders with multi-dimensional attention (2.88M), DiSAN only has 2.35M parameters.

\subsection{Sentiment Analysis} \label{sec:exp_sc}

\begin{table}[htbp] \small
	\centering
	\begin{tabular}{@{}lc@{}}
		\toprule
		\textbf{Model} & \textbf{Test Accu}  \\ \midrule
		MV-RNN \cite{socher2013recursive}&             44.4\\
		RNTN \cite{socher2013recursive}&             45.7\\
		Bi-LSTM \cite{li2015tree}&             49.8\\
		Tree-LSTM \cite{tai2015improved}&             51.0\\
		CNN-non-static \cite{kim2014convolutional}&             48.0\\
		CNN-Tensor \cite{lei2015molding}&             51.2\\
		NCSL \cite{teng2016context}&             51.1\\
		LR-Bi-LSTM \cite{qian2017linguistically}&              50.6\\ \midrule
		Word Embedding with additive attention&  47.47\\ 
		Word Embedding with s2t self-attention&  48.87\\ 
		Multi-head with s2t self-attention&   49.14\\ 
		Bi-LSTM with s2t self-attention&    49.95\\ 
		DiSAN without directions&  49.41\\ \midrule
		DiSAN&                   \textbf{51.72}\\ \bottomrule
	\end{tabular}
	\caption{Test accuracy of fine-grained sentiment analysis on Stanford Sentiment Treebank (SST) dataset.}
	\label{table:sst_exps}
\end{table}

Sentiment analysis aims to analyze the sentiment of a sentence or a paragraph, e.g., a movie or a product review. We use Stanford Sentiment Treebank (SST)\footnote{https://nlp.stanford.edu/sentiment/} \cite{socher2013recursive} for the experiments, and only focus on the fine-grained movie review sentiment classification over five classes, i.e., very negative, negative, neutral, positive and very positive. We use the standard train/dev/test sets split with 8,544/1,101/2,210 samples. Similar to Section \ref{sec:exp_nli}, we employ a single sentence encoding model to obtain a sentence representation $s$ of a movie review, then pass it into a 300D fully connected layer. Finally, a $5$-unit output layer with $\softmax$ is used to calculate a probability distribution over the five classes.

In Table \ref{table:sst_exps}, we compare previous works with DiSAN on test accuracy. To the best of our knowledge, DiSAN improves the last best accuracy (given by CNN-Tensor) by $0.52\%$. Compared to tree-based models with heavy use of the prior structure, e.g., MV-RNN, RNTN and Tree-LSTM, DiSAN outperforms them by $7.32\%$, $6.02\%$ and $0.72\%$, respectively. Additionally, DiSAN achieves better performance than CNN-based models. More recent works tend to focus on lexicon-based sentiment analysis, by exploring sentiment lexicons, negation words and intensity words. Nonetheless, DiSAN still outperforms these fancy models, such as NCSL ($+0.62\%$) and LR-Bi-LSTM ($+1.12\%$).

\begin{figure}[htbp]
	\includegraphics[width=0.45\textwidth]{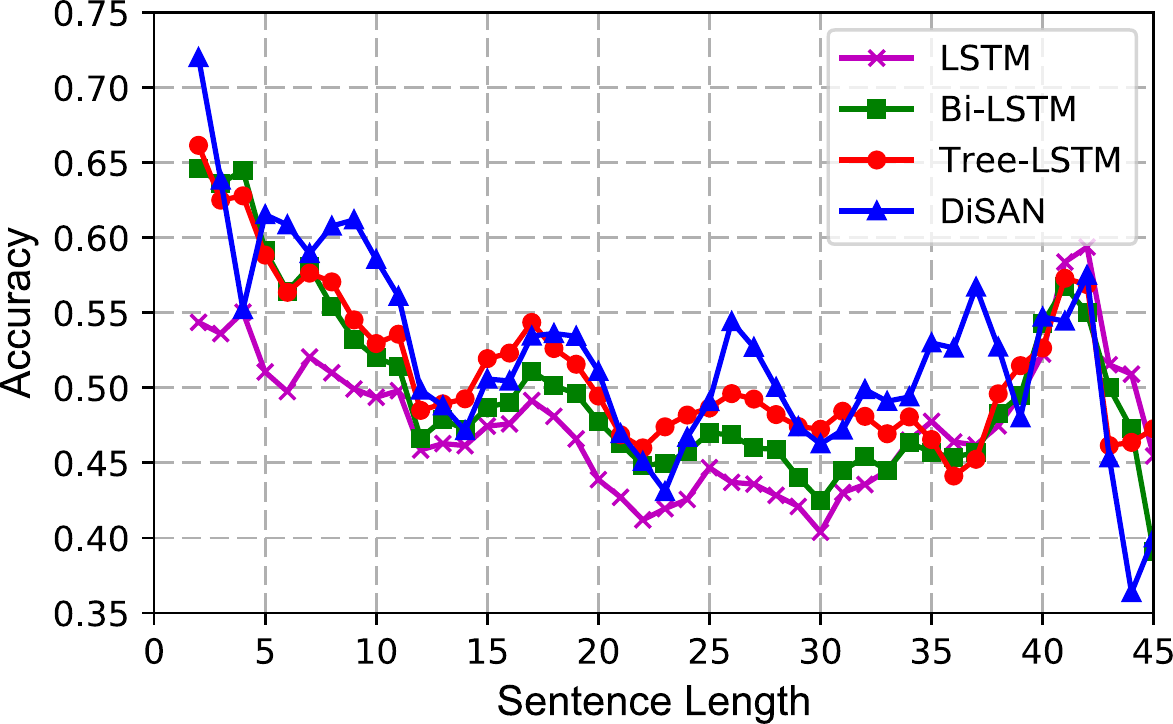}
	\caption{Fine-grained sentiment analysis accuracy vs. sentence length. The results of LSTM, Bi-LSTM and Tree-LSTM are from \citeauthor{tai2015improved} \shortcite{tai2015improved} and the result of DiSAN is the average over five random trials.}
	\label{fig:sst_sent_len_vs_accu} 
	\centering
\end{figure}

It is also interesting to see the performance of different models on the sentences with different lengths. In Figure \ref{fig:sst_sent_len_vs_accu}, we compare LSTM, Bi-LSTM, Tree-LSTM and DiSAN on different sentence lengths. In the range of $(5, 12)$, the length range for most movie review sentences, DiSAN significantly outperforms others. Meanwhile, DiSAN also shows impressive performance for slightly longer sentences or paragraphs in the range of $(25, 38)$. DiSAN performs poorly when the sentence length $\geq 38$, in which however only $3.21\%$ of total movie review sentences lie.

\subsection{Experiments on Other NLP Tasks}

\paragraph{Multi-Genre Natural Language Inference}
Multi-Genre Natural Language Inference (MultiNLI)\footnote{https://www.nyu.edu/projects/bowman/multinli/} \cite{williams2017broad} dataset consists of 433k sentence pairs annotated with textual entailment information. This dataset is similar to SNLI, but it covers more genres of spoken and written text, and supports a distinctive cross-genre generalization evaluation. However, MultiNLI is a quite new dataset, and its leaderboard does not include a session for the sentence-encoding only model. Hence, we only compare DiSAN with the baselines provided at the official website. The results of DiSAN and two sentence-encoding models on the leaderboard are shown in Table \ref{table:multinli_exps}. Note that the prediction accuracies of Matched and Mismatched test datasets are obtained by submitting our test results to Kaggle open evaluation platforms\footnote{https://inclass.kaggle.com/c/multinli-matched-open-evaluation \textit{and} https://inclass.kaggle.com/c/multinli-mismatched-open-evaluation}: \textit{MultiNLI Matched Open Evaluation} and \textit{MultiNLI Mismatched Open Evaluation}.

\begin{table}[htbp]
	\centering
	\begin{tabular}{@{}lcc@{}}
		\toprule
		\textbf{Method}& \textbf{Matched}& \textbf{Mismatched}\\ \midrule
		cBoW&            0.65200& 0.64759\\
		Bi-LSTM&            0.67507& 0.67248\\ \midrule
		DiSAN&            \textbf{0.70977}& \textbf{0.71402}\\ \bottomrule
	\end{tabular}
	\caption{Experimental results of prediction accuracy for different methods on MultiNLI.}
	\label{table:multinli_exps}
\end{table}

\paragraph{Semantic Relatedness}
The task of semantic relatedness aims to predict a similarity degree of a given pair of sentences. We show an experimental comparison of different methods on Sentences Involving Compositional Knowledge (SICK)\footnote{http://clic.cimec.unitn.it/composes/sick.html} dataset \cite{marelli2014sick}. SICK is composed of 9,927 sentence pairs with 4,500/500/4,927 instances for train/dev/test. The regression module on the top of DiSAN is introduced by \citeauthor{tai2015improved} \shortcite{tai2015improved}. The results in Table \ref{table:sick_exps} show that DiSAN outperforms the models from previous works in terms of Pearson's $\bm{r}$ and Spearman's $\bm{\rho}$ indexes.

\begin{table}[htbp] \small
	\centering
	\setlength{\tabcolsep}{1pt}
	\begin{tabular}{@{}lccc@{}}
		\toprule
		\textbf{Model}& \textbf{Pearson's $\bm{r}$} &\textbf{Spearman's $\bm{\rho}$}& \textbf{MSE}  \\ \midrule
		Meaning Factory$^a$&            .8268&               .7721&                .3224\\
		ECNU$^b$&            .8414&               /&                /\\
		DT-RNN$^c$&            .7923 (.0070)&               .7319 (.0071)&               .3822 (.0137) \\ 
		SDT-RNN$^c$&            .7900 (.0042)&               .7304 (.0042)&                .3848 (.0042)\\ 
		Cons. Tree-LSTM$^d$&            .8582 (.0038)&               .7966 (.0053)&                .2734 (.0108)\\ 
		Dep. Tree-LSTM$^d$&            .8676 (.0030)&               .8083 (.0042)&                \textbf{.2532 (.0052)}\\ \midrule
		DiSAN& \textbf{.8695 (.0012)}&  \textbf{.8139 (.0012)}& .2879 (.0036)\\ \bottomrule
	\end{tabular}
	\caption{Experimental results for different methods on SICK sentence relatedness dataset. The reported accuracies are the mean of five runs (standard deviations in parentheses). Cons. and Dep. represent Constituency and Dependency, respectively.  $^a$\cite{bjerva2014meaning}, $^b$\cite{zhao2014ecnu}, $^c$\cite{socher2014grounded}, $^d$\cite{tai2015improved}.}
	\label{table:sick_exps}
\end{table}

\paragraph{Sentence Classifications}

The goal of sentence classification is to correctly predict the class label of a given sentence in various scenarios. We evaluate the models on four sentence classification benchmarks of various NLP tasks, such as sentiment analysis and question-type classification. They are listed as follows. 
1) \textbf{CR}: Customer review \cite{hu2004mining} of various products (cameras, etc.), which is to predict whether the review is  positive or negative; 
2) \textbf{MPQA}: Opinion polarity detection subtask of the MPQA dataset \cite{wiebe2005annotating};
3) \textbf{SUBJ}: Subjectivity dataset \cite{pang2004sentimental} whose labels indicate whether each sentence is subjective or objective;
4) \textbf{TREC}: TREC question-type classification dataset \cite{li2002learning}. The experimental results of DiSAN and existing methods are shown in Table \ref{table:exps_sc_accu}.

\begin{table}[htbp] \small
	\centering
	\setlength{\tabcolsep}{4pt}
	\begin{tabular}{@{}lcccc@{}}
		\toprule
		\textbf{Model}& \textbf{CR} & \textbf{MPQA} & \textbf{SUBJ}& \textbf{TREC}\\ \midrule 
		cBoW$^a$ &	79.9& 86.4& 91.3& 87.3\\
		Skip-thought$^b$ &	81.3& 87.5& 93.6& 92.2\\
		DCNN$^c$&	/& /& /& 93.0\\
		AdaSent$^d$&	83.6 (1.6)& \textbf{90.4 (0.7)}& 92.2 (1.2)& 91.1 (1.0)\\
		SRU$^e$&	\textbf{84.8 (1.3)}& 89.7 (1.1)& 93.4 (0.8)& 93.9 (0.6)\\
		Wide CNNs$^e$&	82.2 (2.2)& 88.8 (1.2)& 92.9 (0.7)& 93.2 (0.5)\\\midrule
		DiSAN &	 \textbf{84.8 (2.0)}&  90.1 (0.4)&  \textbf{94.2 (0.6)}&  \textbf{94.2 (0.1)}\\ \bottomrule
	\end{tabular}
	\caption{Experimental results for different methods on various sentence classification benchmarks. The reported accuracies on CR, MPQA and SUBJ are the mean of 10-fold cross validation, the accuracies on TREC are the mean of dev accuracies of five runs. All standard deviations are in parentheses. $^a$\cite{mikolov2013efficient}, $^b$\cite{kiros2015skip}, $^c$\cite{kalchbrenner2014convolutional}, $^d$\cite{zhao2015self}, $^e$\cite{lei2017sru}.}
	\label{table:exps_sc_accu}
\end{table}

\subsection{Case Study}

\begin{figure}[htbp]
	\centering
	\subfigure[Sentence 1, forward]{
		\label{fig:s1_fw_attn}
		\includegraphics[width=0.22\textwidth]{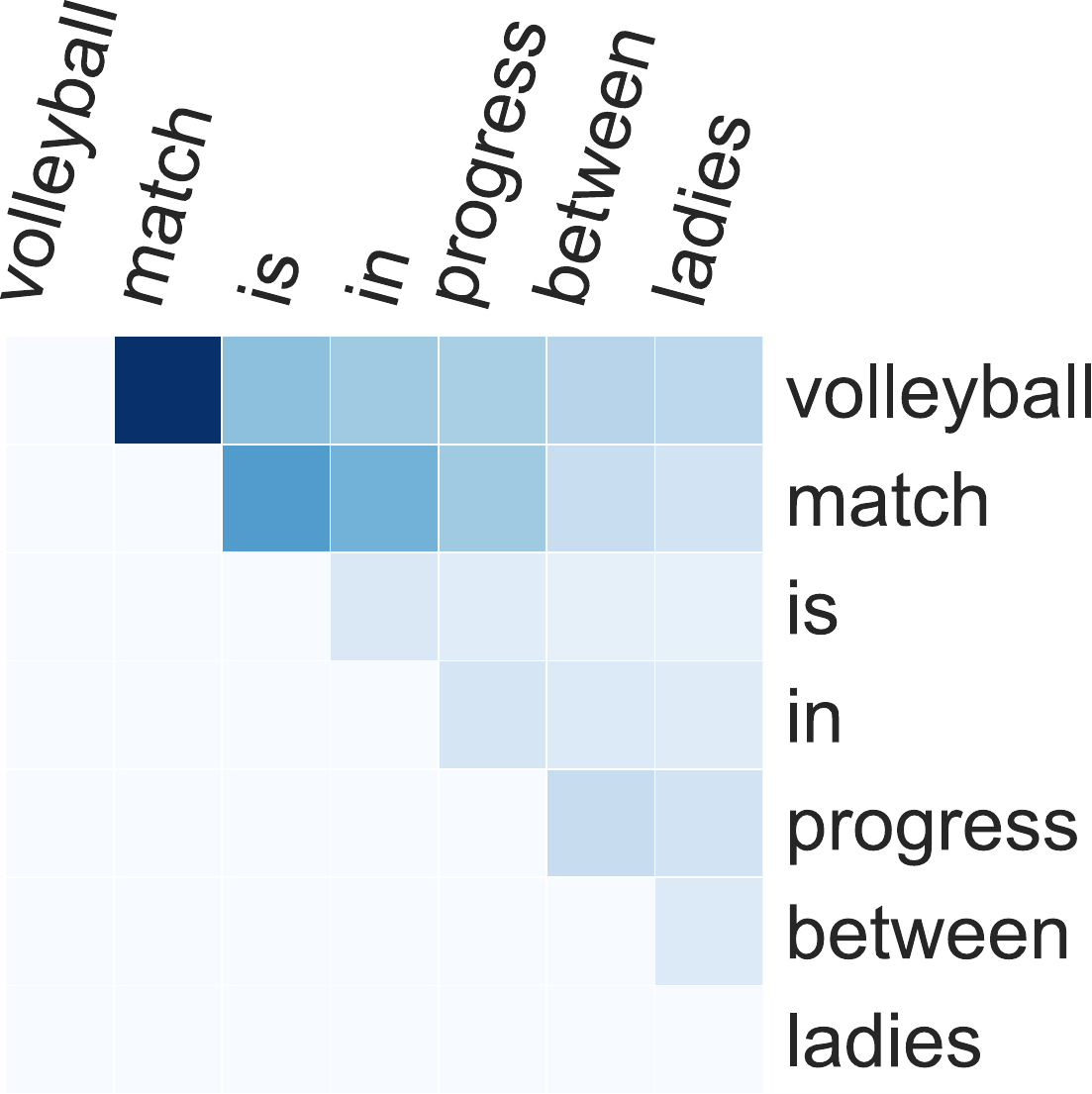}}
	\subfigure[Sentence 1, backward]{
		\label{fig:s1_bw_attn} 
		\includegraphics[width=0.22\textwidth]{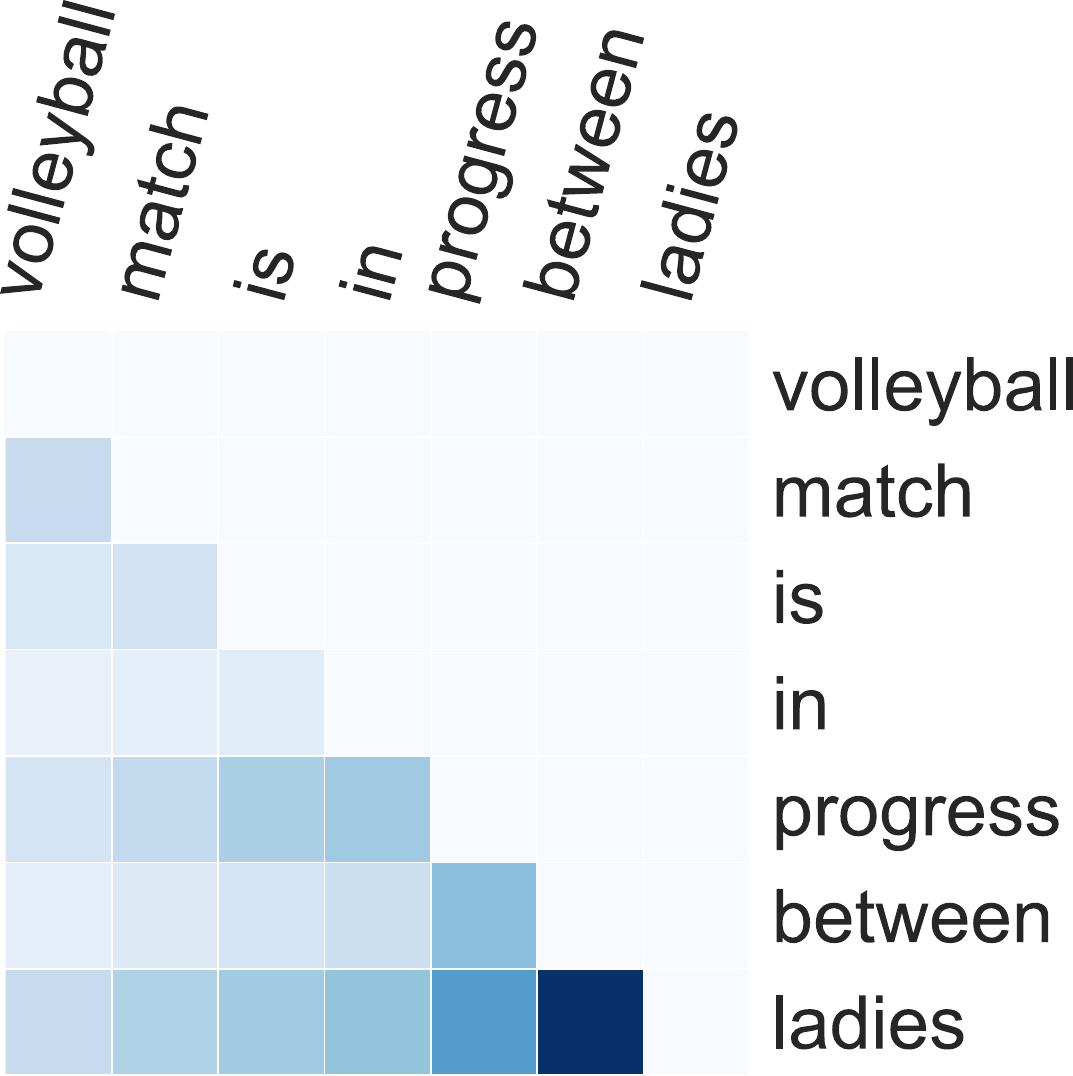}}
	\subfigure[Sentence 2, forward]{
		\label{fig:s2_fw_attn}
		\includegraphics[width=0.22\textwidth]{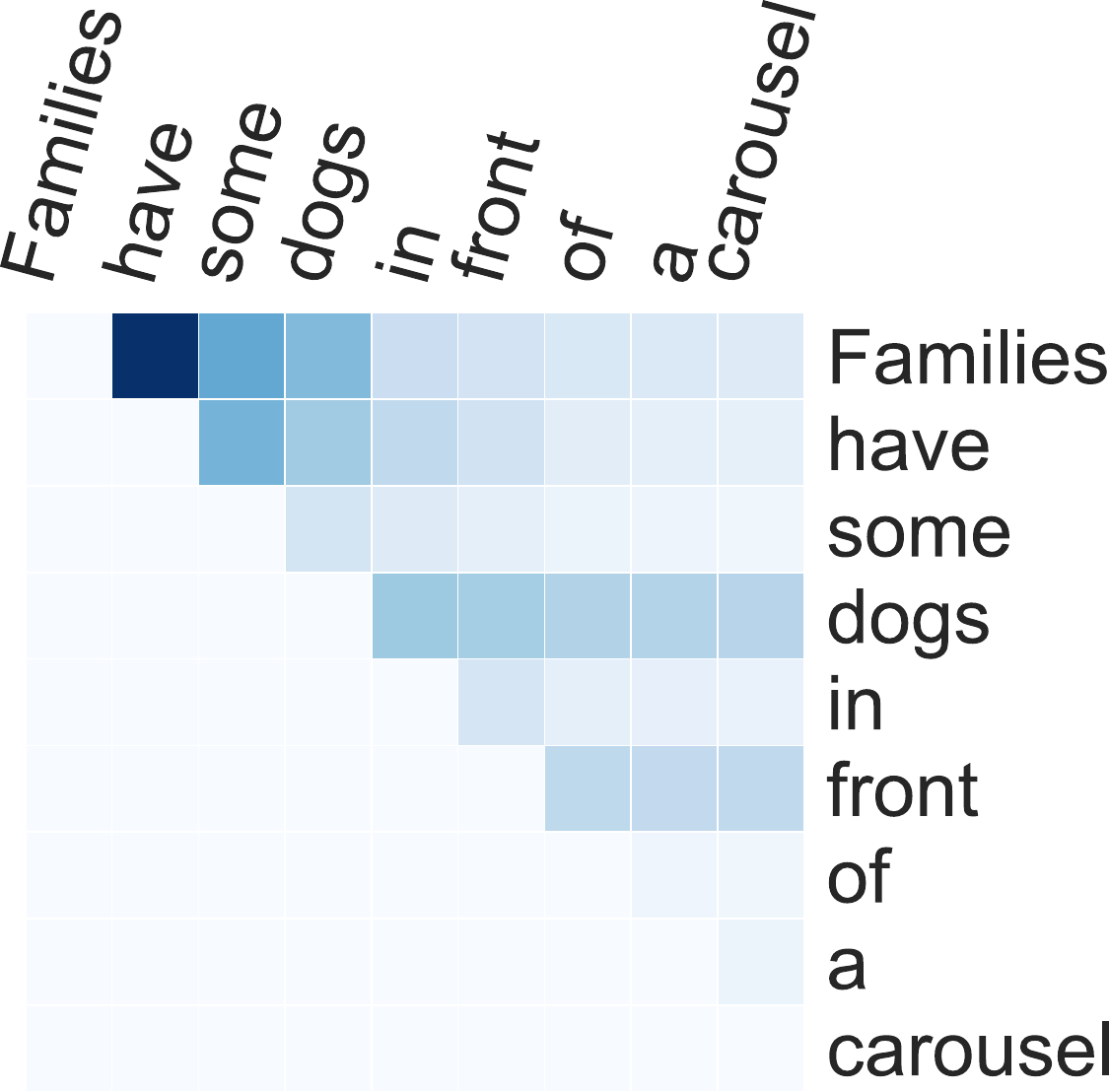}}
	\subfigure[Sentence 2, backward]{
		\label{fig:s2_bw_attn} 
		\includegraphics[width=0.22\textwidth]{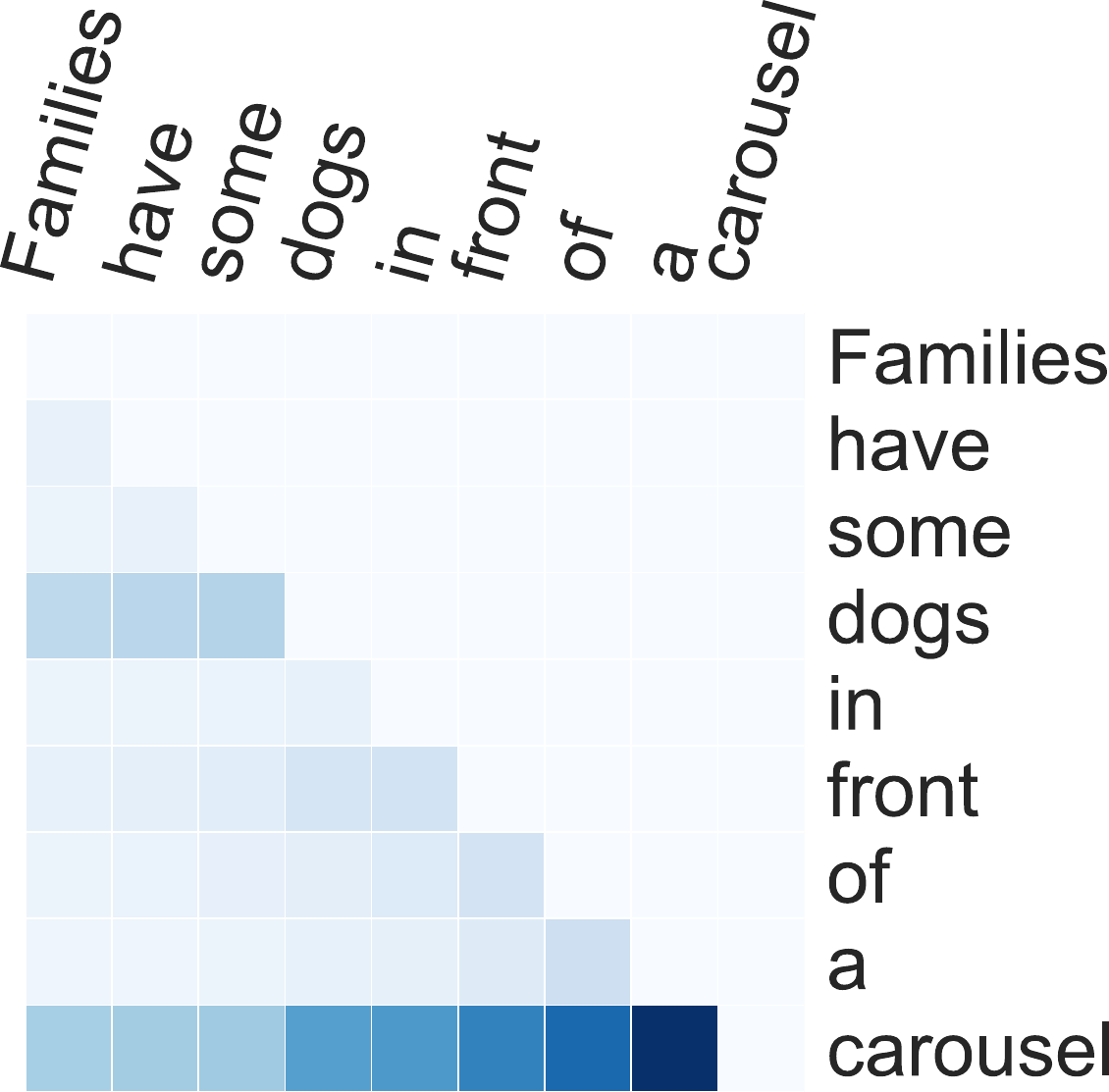}}
	\caption{Attention probability in forward/backward DiSA blocks for the two example sentences.}
	\label{fig:fw_bw_attn}
\end{figure}

\begin{figure}[htbp]
	\centering
	\subfigure[Sentence 1, forward]{
		\label{fig:s1_fw_attn_gate}
		\includegraphics[width=0.22\textwidth]{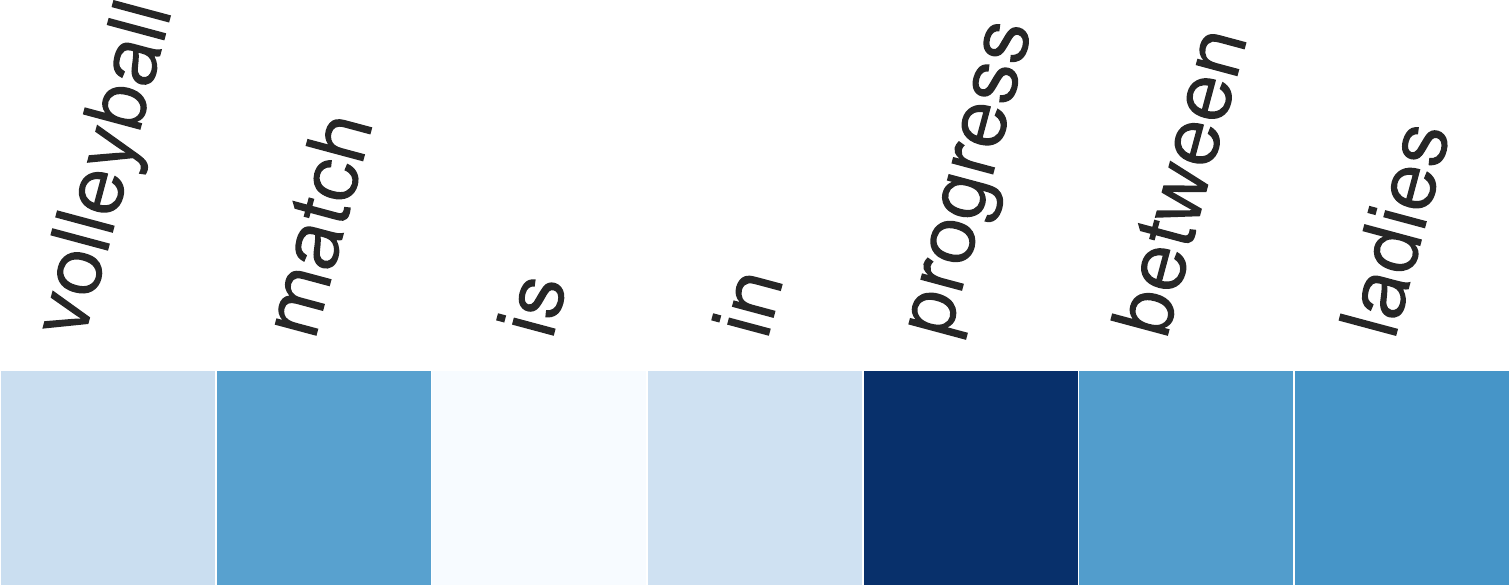}}
	\subfigure[Sentence 1, backward]{
		\label{fig:s1_bw_attn_gate} 
		\includegraphics[width=0.22\textwidth]{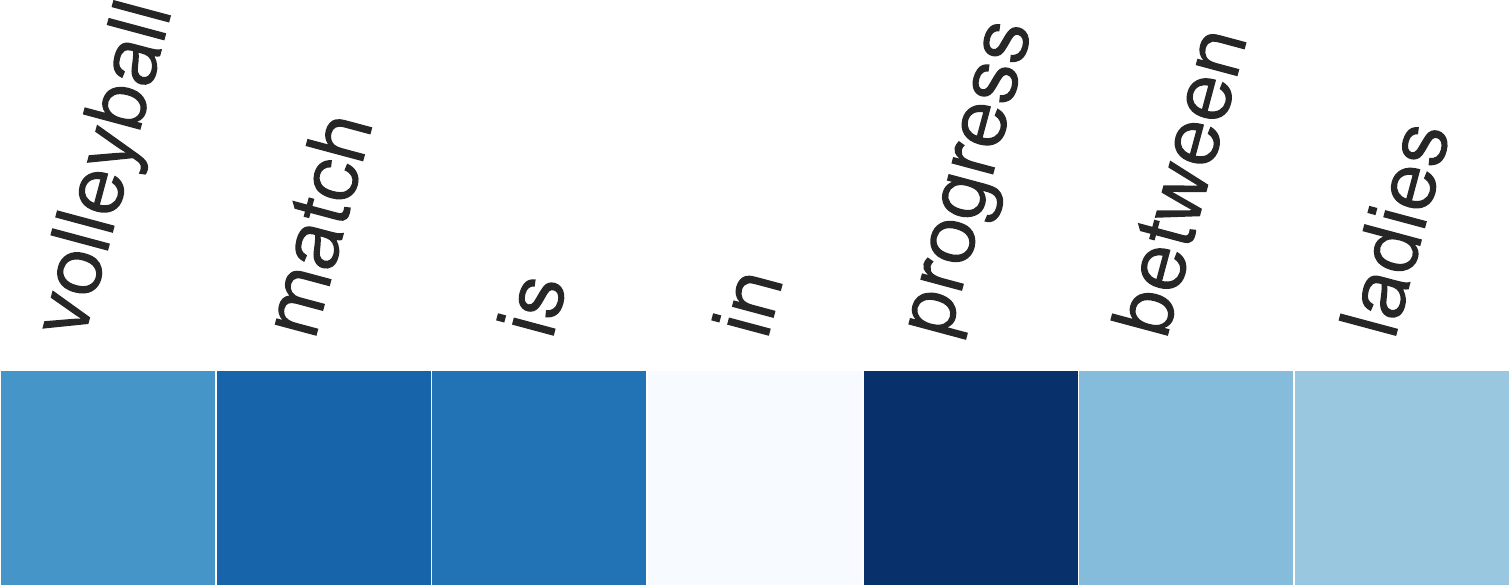}}
	\subfigure[Sentence 2, forward]{
		\label{fig:s2_fw_attn_gate}
		\includegraphics[width=0.22\textwidth]{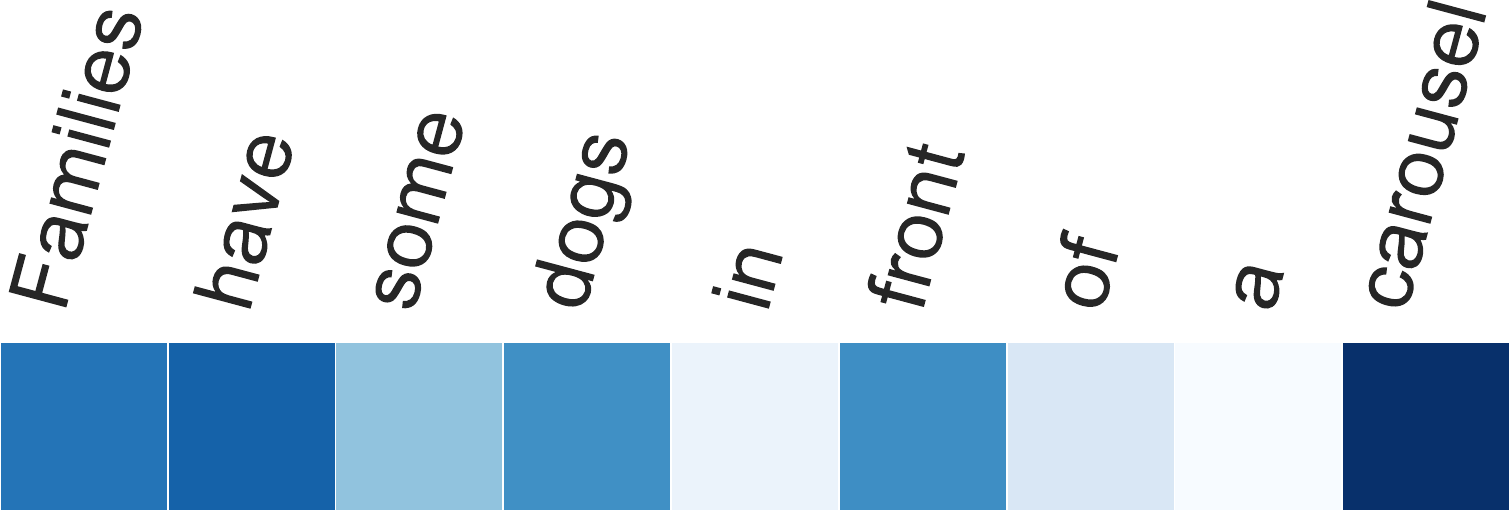}}
	\subfigure[Sentence 2, backward]{
		\label{fig:s2_bw_attn_gate} 
		\includegraphics[width=0.22\textwidth]{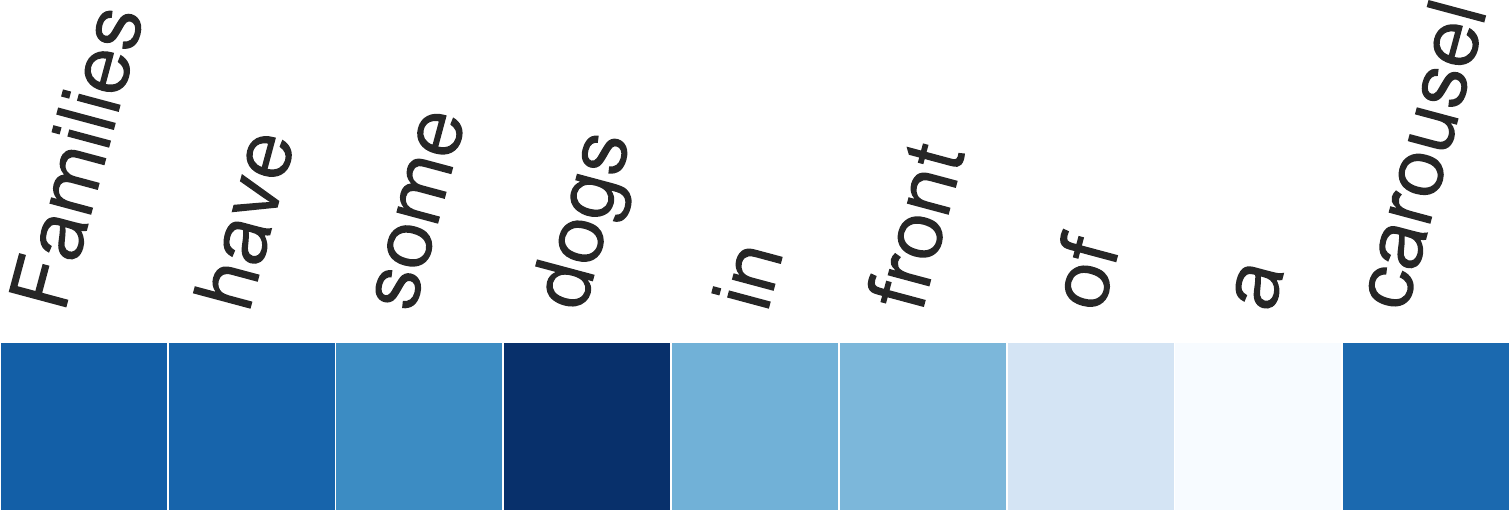}}
	\caption{Fusion Gate $F$ in forward/backward DiSA blocks.}
	\label{fig:fw_bw_attn_gate}
\end{figure}

\begin{figure}[htbp]
	\centering
	\subfigure[\textit{glass} in pair 1]{
		\label{fig:p1_mul_attn}
		\includegraphics[width=0.46\textwidth]{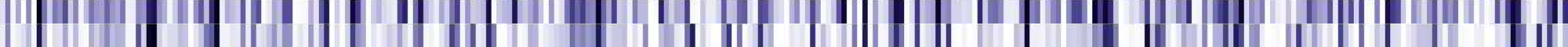}}
	\subfigure[\textit{close} in pair 2]{
		\label{fig:p2_mul_attn}
		\includegraphics[width=0.46\textwidth]{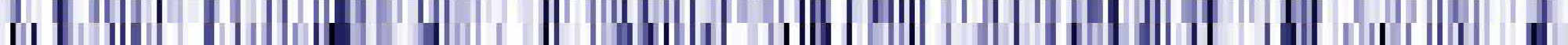}}
	\caption{Two pairs of attention probability comparison of same word in difference sentence contexts.}
	\label{fig:mul_attn}
\end{figure}

To gain a closer view of what dependencies in a sentence can be captured by DiSAN, we visualize the attention probability $p(z=i|\bm{x}, x_j)$ or alignment score by heatmaps. In particular, we will focus primarily on the probability in forward/backward DiSA blocks (Figure \ref{fig:fw_bw_attn}), forward/backward fusion gates $F$ in Eq.(\ref{eq_self_attn_6}) (Figure \ref{fig:fw_bw_attn_gate}), and the probability in multi-dimensional source2token self-attention block (Figure \ref{fig:mul_attn}). For the first two, we desire to demonstrate the dependency at token level, but attention probability in DiSAN is defined on each feature, so we average the probabilities along the feature dimension.

We select two sentences from SNLI test set as examples for this case study. Sentence 1 is \textit{Families have some dogs in front of a carousel} and sentence 2 is \textit{volleyball match is in progress between ladies}.

Figure \ref{fig:fw_bw_attn} shows that1) semantically important words such as nouns and verbs usually get large attention, but stop words (\textit{am, is, are,} etc.) do not; 2) globally important words, e.g., \textit{volleyball, match, ladies} in sentence 1 and \textit{dog, front, carousel} in sentence 2, get large attention from all other words; 3) if a word is important to only some of the other words (e.g. to constitute a phrase or sense-group), it gets large attention only from these words, e.g., attention between \textit{progress, between} in sentence1, and attention between \textit{families, have} in sentence 2.

This also shows that directional information can help to generate context-aware word representation with temporal order encoded. For instance, for word \textit{match} in sentence 1, its forward DiSA focuses more on word \textit{volleyball}, while its backward attention focuses more on \textit{progress} and \textit{ladies}, so the representation of word \textit{match} contains the essential information of the entire sentence, and simultaneously includes the positional order information.

In addition, forward and backward DiSAs can focus on different parts of a sentence. For example, the forward one in sentence 2 pays attention to the word \textit{families}, whereas the backward one focuses on the word \textit{carousel}. Since forward and backward attentions are computed separately, it avoids normalization over multiple significant words to weaken their weights. Note that this is a weakness of traditional attention compared to RNN, especially for long sentences.

In Figure \ref{fig:fw_bw_attn_gate}, we show that the gate value $F$ in Eq.(\ref{eq_self_attn_6}). The gate combines the input and output of masked self-attention. It tends to selects the input representation $\bm{h}$ instead of the output $\bm{s}$ if the corresponding weight in $F$ is large. This shows that the gate values for meaningless words, especially stop words is small. The stop words themselves cannot contribute important information, so only their semantic relations to other words might help to understand the sentence. Hence, the gate tends to use their context features given by masked self-attention.

In Figure \ref{fig:mul_attn}, we show the two multi-dimensional source2token self-attention score vectors of the same word in the two sentences, by their heatmaps. The first pair has two sentences: one is \textit{The \textbf{glass} bottle is big}, and another is \textit{A man is pouring a \textbf{glass} of tea}. They share the same word is \textit{glass} with different meanings. The second pair has two sentences: one is \textit{The restaurant is about to \textbf{close}} and another is \textit{A biker is \textbf{close} to the fountain}. It can be seen that the two attention vectors for the same words are very different due to their different meanings in different contexts. This indicates that the multi-dimensional attention vector is not redundant because it can encode more information than one single score used in traditional attention and it is able to capture subtle difference of the same word in different contexts or sentences. Additionally, it can also alleviate the weakness of the attention over long sequence, which can avoid normalization over entire sequence in traditional attention only once.

\section{Conclusion}
In this paper, we propose two novel attention mechanisms, multi-dimensional attention and directional self-attention. The multi-dimensional attention performs a feature-wise selection over the input sequence for a specific task, and the directional self-attention uses the positional masks to produce the context-aware representations with temporal information encoded. Based on these attentions, Directional Self-Attention Network (DiSAN) is proposed for sentence-encoding without any recurrent or convolutional structure. The experiment results show that DiSAN can achieve state-of-the-art inference quality and outperform existing works (LSTM, etc.) on a wide range of NLP tasks with fewer parameters and higher time efficiency.

In future work, we will explore the approaches to using the proposed attention mechanisms on more sophisticated tasks, e.g. question answering and reading comprehension, to achieve better performance on various benchmarks.

\section{Acknowledgments}
This research was funded by the Australian Government through the Australian Research Council (ARC) under grant 1) LP160100630 partnership with Australia Government Department of Health, and 2) LP150100671 partnership with Australia Research Alliance for Children and Youth (ARACY) and Global Business College Australia (GBCA).

{\fontsize{9.0pt}{10.0pt} \selectfont \bibliography{ref_file}}

\begin{thebibliography}{}

\bibitem[\protect\citeauthoryear{Bahdanau, Cho, and
  Bengio}{2015}]{bahdanau2015neural}
Bahdanau, D.; Cho, K.; and Bengio, Y.
\newblock 2015.
\newblock Neural machine translation by jointly learning to align and
  translate.
\newblock In {\em ICLR}.

\bibitem[\protect\citeauthoryear{Bjerva \bgroup et al\mbox.\egroup
  }{2014}]{bjerva2014meaning}
Bjerva, J.; Bos, J.; Van~der Goot, R.; and Nissim, M.
\newblock 2014.
\newblock The meaning factory: Formal semantics for recognizing textual
  entailment and determining semantic similarity.
\newblock In {\em SemEval@ COLING},  642--646.

\bibitem[\protect\citeauthoryear{Bowman \bgroup et al\mbox.\egroup
  }{2015}]{bowman2015snli}
Bowman, S.~R.; Angeli, G.; Potts, C.; and Manning, C.~D.
\newblock 2015.
\newblock A large annotated corpus for learning natural language inference.
\newblock In {\em EMNLP}.

\bibitem[\protect\citeauthoryear{Bowman \bgroup et al\mbox.\egroup
  }{2016}]{bowman2016fast}
Bowman, S.~R.; Gauthier, J.; Rastogi, A.; Gupta, R.; Manning, C.~D.; and Potts,
  C.
\newblock 2016.
\newblock A fast unified model for parsing and sentence understanding.
\newblock In {\em ACL}.

\bibitem[\protect\citeauthoryear{Chung \bgroup et al\mbox.\egroup
  }{2014}]{chung2014empirical}
Chung, J.; Gulcehre, C.; Cho, K.; and Bengio, Y.
\newblock 2014.
\newblock Empirical evaluation of gated recurrent neural networks on sequence
  modeling.
\newblock In {\em NIPS}.

\bibitem[\protect\citeauthoryear{Clevert, Unterthiner, and
  Hochreiter}{2016}]{clevert2016fast}
Clevert, D.-A.; Unterthiner, T.; and Hochreiter, S.
\newblock 2016.
\newblock Fast and accurate deep network learning by exponential linear units
  (elus).
\newblock In {\em ICLR}.

\bibitem[\protect\citeauthoryear{Glorot and
  Bengio}{2010}]{glorot2010understanding}
Glorot, X., and Bengio, Y.
\newblock 2010.
\newblock Understanding the difficulty of training deep feedforward neural
  networks.
\newblock In {\em Proceedings of the Thirteenth International Conference on
  Artificial Intelligence and Statistics},  249--256.

\bibitem[\protect\citeauthoryear{Graves, Jaitly, and
  Mohamed}{2013}]{graves2013hybrid}
Graves, A.; Jaitly, N.; and Mohamed, A.-r.
\newblock 2013.
\newblock Hybrid speech recognition with deep bidirectional lstm.
\newblock In {\em Automatic Speech Recognition and Understanding (ASRU), 2013
  IEEE Workshop on},  273--278.
\newblock IEEE.

\bibitem[\protect\citeauthoryear{Hermann \bgroup et al\mbox.\egroup
  }{2015}]{hermann2015teaching}
Hermann, K.~M.; Kocisky, T.; Grefenstette, E.; Espeholt, L.; Kay, W.; Suleyman,
  M.; and Blunsom, P.
\newblock 2015.
\newblock Teaching machines to read and comprehend.
\newblock In {\em NIPS}.

\bibitem[\protect\citeauthoryear{Hochreiter and
  Schmidhuber}{1997}]{hochreiter1997long}
Hochreiter, S., and Schmidhuber, J.
\newblock 1997.
\newblock Long short-term memory.
\newblock {\em Neural computation} 9(8):1735--1780.

\bibitem[\protect\citeauthoryear{Hu and Liu}{2004}]{hu2004mining}
Hu, M., and Liu, B.
\newblock 2004.
\newblock Mining and summarizing customer reviews.
\newblock In {\em Proceedings of the tenth ACM SIGKDD international conference
  on Knowledge discovery and data mining},  168--177.
\newblock ACM.

\bibitem[\protect\citeauthoryear{Hu, Peng, and Qiu}{2017}]{hu2017mnemonic}
Hu, M.; Peng, Y.; and Qiu, X.
\newblock 2017.
\newblock Reinforced mnemonic reader for machine comprehension.
\newblock {\em arXiv preprint arXiv:1705.02798}.

\bibitem[\protect\citeauthoryear{Kalchbrenner, Grefenstette, and
  Blunsom}{2014}]{kalchbrenner2014convolutional}
Kalchbrenner, N.; Grefenstette, E.; and Blunsom, P.
\newblock 2014.
\newblock A convolutional neural network for modelling sentences.
\newblock {\em arXiv preprint arXiv:1404.2188}.

\bibitem[\protect\citeauthoryear{Kim \bgroup et al\mbox.\egroup
  }{2016}]{kim2016character}
Kim, Y.; Jernite, Y.; Sontag, D.; and Rush, A.~M.
\newblock 2016.
\newblock Character-aware neural language models.
\newblock In {\em AAAI}.

\bibitem[\protect\citeauthoryear{Kim}{2014}]{kim2014convolutional}
Kim, Y.
\newblock 2014.
\newblock Convolutional neural networks for sentence classification.
\newblock In {\em EMNLP}.

\bibitem[\protect\citeauthoryear{Kingma and Ba}{2015}]{kingma2014adam}
Kingma, D., and Ba, J.
\newblock 2015.
\newblock Adam: A method for stochastic optimization.
\newblock In {\em ICLR}.

\bibitem[\protect\citeauthoryear{Kiros \bgroup et al\mbox.\egroup
  }{2015}]{kiros2015skip}
Kiros, R.; Zhu, Y.; Salakhutdinov, R.~R.; Zemel, R.; Urtasun, R.; Torralba, A.;
  and Fidler, S.
\newblock 2015.
\newblock Skip-thought vectors.
\newblock In {\em NIPS}.

\bibitem[\protect\citeauthoryear{Kokkinos and
  Potamianos}{2017}]{kokkinos2017structural}
Kokkinos, F., and Potamianos, A.
\newblock 2017.
\newblock Structural attention neural networks for improved sentiment analysis.
\newblock {\em arXiv preprint arXiv:1701.01811}.

\bibitem[\protect\citeauthoryear{Lei and Zhang}{2017}]{lei2017sru}
Lei, T., and Zhang, Y.
\newblock 2017.
\newblock Training rnns as fast as cnns.
\newblock {\em arXiv preprint arXiv:1709.02755}.

\bibitem[\protect\citeauthoryear{Lei, Barzilay, and
  Jaakkola}{2015}]{lei2015molding}
Lei, T.; Barzilay, R.; and Jaakkola, T.
\newblock 2015.
\newblock Molding cnns for text: non-linear, non-consecutive convolutions.
\newblock In {\em EMNLP}.

\bibitem[\protect\citeauthoryear{Li and Roth}{2002}]{li2002learning}
Li, X., and Roth, D.
\newblock 2002.
\newblock Learning question classifiers.
\newblock In {\em ACL}.

\bibitem[\protect\citeauthoryear{Li \bgroup et al\mbox.\egroup
  }{2015}]{li2015tree}
Li, J.; Luong, M.-T.; Jurafsky, D.; and Hovy, E.
\newblock 2015.
\newblock When are tree structures necessary for deep learning of
  representations?
\newblock {\em arXiv preprint arXiv:1503.00185}.

\bibitem[\protect\citeauthoryear{Liu \bgroup et al\mbox.\egroup
  }{2016}]{liu2016learning}
Liu, Y.; Sun, C.; Lin, L.; and Wang, X.
\newblock 2016.
\newblock Learning natural language inference using bidirectional lstm model
  and inner-attention.
\newblock {\em arXiv preprint arXiv:1605.09090}.

\bibitem[\protect\citeauthoryear{Luong, Pham, and
  Manning}{2015}]{luong2015effective}
Luong, M.-T.; Pham, H.; and Manning, C.~D.
\newblock 2015.
\newblock Effective approaches to attention-based neural machine translation.
\newblock In {\em EMNLP}.

\bibitem[\protect\citeauthoryear{Marelli \bgroup et al\mbox.\egroup
  }{2014}]{marelli2014sick}
Marelli, M.; Menini, S.; Baroni, M.; Bentivogli, L.; Bernardi, R.; and
  Zamparelli, R.
\newblock 2014.
\newblock A sick cure for the evaluation of compositional distributional
  semantic models.
\newblock In {\em LREC}.

\bibitem[\protect\citeauthoryear{Mikolov \bgroup et al\mbox.\egroup
  }{2013a}]{mikolov2013efficient}
Mikolov, T.; Chen, K.; Corrado, G.; and Dean, J.
\newblock 2013a.
\newblock Efficient estimation of word representations in vector space.
\newblock {\em arXiv preprint arXiv:1301.3781}.

\bibitem[\protect\citeauthoryear{Mikolov \bgroup et al\mbox.\egroup
  }{2013b}]{mikolov2013distributed}
Mikolov, T.; Sutskever, I.; Chen, K.; Corrado, G.~S.; and Dean, J.
\newblock 2013b.
\newblock Distributed representations of words and phrases and their
  compositionality.
\newblock In {\em NIPS}.

\bibitem[\protect\citeauthoryear{Mou \bgroup et al\mbox.\egroup
  }{2016}]{mou2016natural}
Mou, L.; Men, R.; Li, G.; Xu, Y.; Zhang, L.; Yan, R.; and Jin, Z.
\newblock 2016.
\newblock Natural language inference by tree-based convolution and heuristic
  matching.
\newblock In {\em ACL}.

\bibitem[\protect\citeauthoryear{Munkhdalai and
  Yu}{2017a}]{munkhdalai2016neural_2}
Munkhdalai, T., and Yu, H.
\newblock 2017a.
\newblock Neural semantic encoders.
\newblock In {\em EACL}.

\bibitem[\protect\citeauthoryear{Munkhdalai and
  Yu}{2017b}]{munkhdalai2017neural}
Munkhdalai, T., and Yu, H.
\newblock 2017b.
\newblock Neural tree indexers for text understanding.
\newblock In {\em EACL}.

\bibitem[\protect\citeauthoryear{Pang and Lee}{2004}]{pang2004sentimental}
Pang, B., and Lee, L.
\newblock 2004.
\newblock A sentimental education: Sentiment analysis using subjectivity
  summarization based on minimum cuts.
\newblock In {\em ACL}.

\bibitem[\protect\citeauthoryear{Pennington, Socher, and
  Manning}{2014}]{pennington2014glove}
Pennington, J.; Socher, R.; and Manning, C.~D.
\newblock 2014.
\newblock Glove: Global vectors for word representation.
\newblock In {\em EMNLP}.

\bibitem[\protect\citeauthoryear{Qian, Huang, and
  Zhu}{2017}]{qian2017linguistically}
Qian, Q.; Huang, M.; and Zhu, X.
\newblock 2017.
\newblock Linguistically regularized lstms for sentiment classification.
\newblock In {\em ACL}.

\bibitem[\protect\citeauthoryear{Rush, Chopra, and
  Weston}{2015}]{rush2015neural}
Rush, A.~M.; Chopra, S.; and Weston, J.
\newblock 2015.
\newblock A neural attention model for abstractive sentence summarization.
\newblock In {\em EMNLP}.

\bibitem[\protect\citeauthoryear{Seo \bgroup et al\mbox.\egroup
  }{2017}]{seo2017bidirectional}
Seo, M.; Kembhavi, A.; Farhadi, A.; and Hajishirzi, H.
\newblock 2017.
\newblock Bidirectional attention flow for machine comprehension.
\newblock In {\em ICLR}.

\bibitem[\protect\citeauthoryear{Shang, Lu, and Li}{2015}]{shang2015neural}
Shang, L.; Lu, Z.; and Li, H.
\newblock 2015.
\newblock Neural responding machine for short-text conversation.
\newblock In {\em ACL}.

\bibitem[\protect\citeauthoryear{Socher \bgroup et al\mbox.\egroup
  }{2013}]{socher2013recursive}
Socher, R.; Perelygin, A.; Wu, J.~Y.; Chuang, J.; Manning, C.~D.; Ng, A.~Y.;
  Potts, C.; et~al.
\newblock 2013.
\newblock Recursive deep models for semantic compositionality over a sentiment
  treebank.
\newblock In {\em EMNLP}.

\bibitem[\protect\citeauthoryear{Socher \bgroup et al\mbox.\egroup
  }{2014}]{socher2014grounded}
Socher, R.; Karpathy, A.; Le, Q.~V.; Manning, C.~D.; and Ng, A.~Y.
\newblock 2014.
\newblock Grounded compositional semantics for finding and describing images
  with sentences.
\newblock {\em Transactions of the Association for Computational Linguistics}
  2:207--218.

\bibitem[\protect\citeauthoryear{Srivastava \bgroup et al\mbox.\egroup
  }{2014}]{srivastava2014dropout}
Srivastava, N.; Hinton, G.~E.; Krizhevsky, A.; Sutskever, I.; and
  Salakhutdinov, R.
\newblock 2014.
\newblock Dropout: a simple way to prevent neural networks from overfitting.
\newblock {\em Journal of Machine Learning Research} 15(1):1929--1958.

\bibitem[\protect\citeauthoryear{Sukhbaatar \bgroup et al\mbox.\egroup
  }{2015}]{sukhbaatar2015end}
Sukhbaatar, S.; Weston, J.; Fergus, R.; et~al.
\newblock 2015.
\newblock End-to-end memory networks.
\newblock In {\em NIPS}.

\bibitem[\protect\citeauthoryear{Tai, Socher, and
  Manning}{2015}]{tai2015improved}
Tai, K.~S.; Socher, R.; and Manning, C.~D.
\newblock 2015.
\newblock Improved semantic representations from tree-structured long
  short-term memory networks.
\newblock In {\em ACL}.

\bibitem[\protect\citeauthoryear{Teng, Vo, and Zhang}{2016}]{teng2016context}
Teng, Z.; Vo, D.-T.; and Zhang, Y.
\newblock 2016.
\newblock Context-sensitive lexicon features for neural sentiment analysis.
\newblock In {\em EMNLP}.

\bibitem[\protect\citeauthoryear{Vaswani \bgroup et al\mbox.\egroup
  }{2017}]{vaswani2017attention}
Vaswani, A.; Shazeer; Noam; Parmar, N.; Uszkoreit, J.; Jones, L.; Gomez, A.~N.;
  Kaiser, L.; and Polosukhin, I.
\newblock 2017.
\newblock Attention is all you need.
\newblock In {\em NIPS}.

\bibitem[\protect\citeauthoryear{Vendrov \bgroup et al\mbox.\egroup
  }{2016}]{vendrov2016order}
Vendrov, I.; Kiros, R.; Fidler, S.; and Urtasun, R.
\newblock 2016.
\newblock Order-embeddings of images and language.
\newblock In {\em ICLR}.

\bibitem[\protect\citeauthoryear{Wiebe, Wilson, and
  Cardie}{2005}]{wiebe2005annotating}
Wiebe, J.; Wilson, T.; and Cardie, C.
\newblock 2005.
\newblock Annotating expressions of opinions and emotions in language.
\newblock {\em Language resources and evaluation} 39(2):165--210.

\bibitem[\protect\citeauthoryear{Williams, Nangia, and
  Bowman}{2017}]{williams2017broad}
Williams, A.; Nangia, N.; and Bowman, S.~R.
\newblock 2017.
\newblock A broad-coverage challenge corpus for sentence understanding through
  inference.
\newblock {\em arXiv preprint arXiv:1704.05426}.

\bibitem[\protect\citeauthoryear{Zeiler}{2012}]{zeiler2012adadelta}
Zeiler, M.~D.
\newblock 2012.
\newblock Adadelta: an adaptive learning rate method.
\newblock {\em arXiv preprint arXiv:1212.5701}.

\bibitem[\protect\citeauthoryear{Zhao, Lu, and Poupart}{2015}]{zhao2015self}
Zhao, H.; Lu, Z.; and Poupart, P.
\newblock 2015.
\newblock Self-adaptive hierarchical sentence model.
\newblock In {\em IJCAI}.

\bibitem[\protect\citeauthoryear{Zhao, Zhu, and Lan}{2014}]{zhao2014ecnu}
Zhao, J.; Zhu, T.; and Lan, M.
\newblock 2014.
\newblock Ecnu: One stone two birds: Ensemble of heterogenous measures for
  semantic relatedness and textual entailment.
\newblock In {\em SemEval@ COLING},  271--277.

\end{thebibliography}
\bibliographystyle{aaai}
\end{document}